\documentclass{ametsocV6.1}

\usepackage{booktabs}
\usepackage{siunitx}
\usepackage{multirow}
\usepackage{bbm}
\usepackage[hidelinks]{hyperref}
\usepackage{cleveref}

\makeatletter
\newcounter{appx}
\renewcommand{\theappx}{\Alph{appx}}
\crefname{appx}{appendix}{appendices}
\Crefname{appx}{Appendix}{Appendices}
\newcommand{\amsappendix}[2]{%
  \refstepcounter{appx}%
  \protected@edef\@tmpappx{\theappx}%
  \expandafter\appendix\expandafter[\@tmpappx]%
  \appendixtitle{#2}%
  \label{#1}%
}
\makeatother

\makeatletter
\renewcommand\p@subsection{\thesection}
\renewcommand\p@subsubsection{\thesection\thesubsection}
\makeatother

\crefname{figure}{Fig.}{Figs.}
\Crefname{figure}{Figure}{Figures}
\crefname{table}{Table}{Tables}
\Crefname{table}{Table}{Tables}
\crefname{equation}{Eq.}{Eqs.}
\Crefname{equation}{Equation}{Equations}
\crefname{section}{section}{sections}
\Crefname{section}{Section}{Sections}

\newcommand{\tessera}{\textsc{Tessera}}
\newcommand{\convcnp}{ConvCNP}
\newcommand{\xstar}{x_\star}
\newcommand{\ystar}{y_\star}

\title{Earth observation embeddings are effective subgrid descriptors for probabilistic weather downscaling}

\authors{Pedro Sousa,\aff{a}\correspondingauthor{Pedro Sousa, pmms2@cam.ac.uk}
Will Tebbutt,\aff{b}
Sadiq Jaffer,\aff{a}
Robin Young,\aff{a}
Anil Madhavapeddy,\aff{a}
and Richard E. Turner\aff{b}}

\affiliation{\aff{a}{Department of Computer Science and Technology, University of Cambridge, United Kingdom}\\
\aff{b}{Department of Engineering, University of Cambridge, United Kingdom}}

\abstract{
    Global weather reanalyses and forecasts resolve the evolving atmospheric state on coarse grids, but site-specific applications require predictions at arbitrary locations where near-surface conditions also depend on unresolved terrain and land-surface properties. Existing probabilistic downscalers address this gap using hand-crafted topographic and surface descriptors. We ask instead whether Earth observation foundation models can provide transferable subgrid surface representations for probabilistic weather downscaling.
    We augment a convolutional conditional neural process (ConvCNP) that downscales coarse ERA5 fields at ${\sim}25$\,km resolution with a learned local surface descriptor, obtained by compressing a patch of \tessera{} embeddings at $10$\,m resolution. Although these embeddings summarize annual surface conditions, they improve downscaling by encoding persistent surface properties that capture a location's departure from the coarse-grid atmospheric state. Across five climatically diverse regions, the embedding improves point and probabilistic skill at stations held out in both space and time, overall improving CRPS skill by $11.5\%$ for instantaneous $2$\,m temperature and $6.2\%$ for $10$\,m wind speed relative to a topography-only ConvCNP baseline. A hand-crafted descriptor incorporating richer surface information than topography alone captures comparable persistent subgrid signal but yields far smaller predictive gains than the learned embedding representation.
    These improvements persist when forecasts from the Aurora AI model replace ERA5 reanalysis fields and when predicting at newly deployed weather station networks. To our knowledge, this is the first evidence that long-timescale Earth observation embeddings can support short-timescale weather downscaling where subgrid departures are systematically structured by persistent surface properties.
}

\begin{document}
\nolinenumbers
\let\internallinenumbers\relax
\maketitle

\statement
Weather forecasts and reanalyses are produced on grids tens of kilometers across, but decisions are made at specific locations, such as wind farms or agricultural fields, where conditions can differ sharply from the grid-cell average. Bridging that gap has usually relied on elevation-based descriptions of local terrain. We show that a general-purpose description of the land surface, learned from satellite imagery and available anywhere on land, performs significantly better. The approach improves local temperature and wind speed predictions at weather stations the model has never seen, including in regions where observations are scarce. This provides a practical bridge between global weather models and advances in Earth observation foundation models, enabling more accurate predictions at specific sites.

\section{Introduction}
\label{sec:intro}

Weather reanalyses and forecasts are often produced on relatively coarse spatial grids (e.g., $25$\,km), struggling to represent weather variability at the local scales required by many downstream applications~\citep{rampal2024,yoshimura2008}. The near-surface state at specific locations can depart substantially from the cell average captured by a coarse grid because a range of subgrid processes produces structured variation within individual cells~\citep{marinescu2024}.

This subgrid variation reflects both local surface properties and interactions with the surrounding environment. Land cover affects near-surface temperature within a single coarse cell: urban--rural contrast associated with the urban heat island is typically $1$--$3\,^\circ$C hotter~\citep{akbari2001} and larger on clear, calm nights~\citep{oke1982}. Terrain also shapes temperature through processes such as cold-air drainage, whereby dense air pools in valley floors and hollows can sit $4$--$8\,^\circ$C colder than slopes only tens of metres higher at the same coarse-grid elevation~\citep{coldair}.

Similarly, wind accelerates when channeled along valleys or through terrain gaps and slows over forest canopy or built surfaces relative to open ground and water. Recent kilometre-scale wind speed forecasts over Texas have improved near-surface wind speed predictions by ${\sim}23\%$ over ECMWF's operational global forecasts, with corresponding gains for wind-power forecasting~\citep{jin2025}. These effects occur below the resolution of a coarse atmospheric grid and are properties of \emph{where a point sits on the surface} rather than of the large-scale flow alone.

Statistical downscaling is the standard approach for capturing such subgrid variation by learning relationships between coarse-scale atmospheric predictors and local observations. This includes perfect prognosis (PP), where models are trained on analyzed large-scale fields such as reanalysis products, and model output statistics (MOS), trained directly on model forecast outputs~\citep{maraun2018}. The common aim is to recover the local value at a target location from the resolved large-scale state, together with information about the unresolved local environment. Generalizing to \emph{unseen locations at unseen times} holds the most operational value and is this paper's central task, supporting newly instrumented sites or locations that will never host a sensor.

An alternative to off-grid downscaling is grid-to-grid super-resolution modeling that maps a coarse field to a finer one, thereby avoiding an explicit off-grid mapping if the predicted field's resolution is sufficient. Examples include convolutional regression~\citep{banomedina2020,hohlein2020} and generative models that sample a fine-scale field: GANs for precipitation~\citep{harris2022} and for wind over complex terrain~\citep{miralles2022}, and diffusion models reaching kilometre scale~\citep{corrdiff}. However, depending on the downstream task's required resolution, the output grid might itself carry subgrid variation. Additionally, grid-to-grid methods require dense, high-resolution target fields for training, often from regional NWP models, so the learned downscaler may inherit their biases and structural assumptions.

Closer to our setting, several station-level forecasting and downscaling methods improve local predictions by conditioning on observations from the target station or incorporating station measurements into a high-resolution prediction framework~\citep{yang2024,metnet3}. These, by construction, fit only locations already served with weather stations. In contrast, we deliberately withhold local history, requiring the downscaling model to predict at previously unobserved sites.

Generalizing beyond instrumented sites means predicting directly at arbitrary target locations, relying only on coarse atmospheric fields and local surface descriptors, while being trained against ground-truth station observations. Although this replaces dense gridded supervision with sparse, scattered point targets, it defines the downscaling objective directly against station measurements rather than model-derived high-resolution fields, consistent with a recent line of observation-driven forecasting~\citep{aardvark, aidop}.

Off-grid, multi-site, probabilistic prediction from a gridded field is precisely what the convolutional conditional neural process (\convcnp{})~\citep{gordon2019}, a translation-equivariant member of the conditional neural process family~\citep{garnelo2018}, is designed for. \citet{vaughan2021} applied a \convcnp{} to map a coarse reanalysis grid to a predictive distribution at arbitrary station locations, trained by maximum likelihood learning on station data. The same mechanism was adopted for the downscaling stage of ~\citet{aardvark}, one of the first end-to-end data-driven global forecasting models. 

In \citet{vaughan2021}'s \convcnp{}, subgrid signal enters through a topographic descriptor specific to the station via three scalars summarizing elevation and terrain prominence. Terrain, however, is only one axis of the near-surface state, as land cover, canopy structure, surface roughness, soil moisture, water and built structure all modulate how a point departs from its coarse-cell mean. \citet{bakketun2026} explores capturing this broader near-surface state through a hand-crafted enumeration of surface descriptors in a separate downscaling framework. 

Rather than specifying these surface properties individually, we ask whether Earth observation (EO) geospatial foundation models can provide a learned subgrid surface representation for off-grid downscaling. \tessera{}~\citep{tessera2025} is trained self-supervised on a full year of paired Sentinel-1 radar and Sentinel-2 optical imagery and emits a $128$-dimensional embedding for every $10$\,m pixel, summarizing annual surface characteristics without requiring them to be enumerated in advance. It is precomputed and globally available at any terrestrial query location, including sites that have never hosted an instrument. EO embeddings of this kind have been applied mainly to land-cover, crop and change-detection mapping; we are not aware of prior use as a subgrid descriptor for near-surface weather.

This work's main contributions are as follows:
\begin{enumerate}
    \item \textbf{Construct a learned subgrid descriptor from EO foundation-model embeddings.} A single \tessera{} pixel is too local to capture the range of near-surface properties affecting a weather-station observation, so we pre-train a VAE to compress a neighbourhood of per-pixel embeddings into a compact local surface descriptor (\cref{sec:method-tessera}).

    \item \textbf{Demonstrate consistent gains in off-grid probabilistic downscaling.} Conditioning \convcnp{}'s decoder on \tessera{} alongside topography improves point (MAE and RMSE) and probabilistic skill (CRPS) across five climatically diverse regions for $2$\,m temperature and $10$\,m wind speed. Ablating for the impact of learned representation, a control with richer hand-crafted surface features improves over topography alone but recovers only a minority of the embedding's gain (\cref{sec:skill}).

    \item \textbf{Characterize what the learned surface representation contributes.} \tessera{}'s contribution differs by variable: for $2$\,m temperature, most fine-scale structure is expressible through elevation, so the embedding acts primarily as a transferable land-surface prior where observations are spatially scarce. For $10$\,m wind speed, persistent residual structure is better organized by richer surface representations, and \tessera{} induces substantially more fine-scale structure in downscaled weather fields (\cref{sec:residual}, \cref{sec:maps}).

    \item \textbf{Demonstrate the method's robustness to forecast weather fields.} Replacing ERA5 with Aurora forecast fields~\citep{aurora}, the embedding's benefit persists across lead times, demonstrating its value in an operationally realistic weather forecasting deployment (\cref{sec:aurora}).

    \item \textbf{Show the embedding improves sample efficiency.} In a simulated Norwegian station network ramp-up, \tessera{} provides competitive wind-speed downscaling before any local observations are available, outperforming interpolated ERA5 at initial deployment; the topography-only baseline fails to reach the same error level even after 6 years of accumulated local data. This shows the data efficiency reported for \tessera{} upstream~\citep{tessera2025} carries through to weather downscaling.
\end{enumerate}

\section{Data and methods}
\label{sec:method}

We describe the coarse atmospheric inputs, station observations, and per-location surface descriptors used by the model. We then introduce the \convcnp{} downscaling framework, including how the \tessera{} surface descriptor is constructed at arbitrary target locations, before detailing our experimental setup. We recap only the \convcnp{} components needed for the present method and refer to \citet{vaughan2021} for the full architecture; additional implementation details are provided in \cref{app:impl}.

\subsection{Data}
\label{sec:method-data}

\paragraph*{Coarse weather grid.}
We downscale ERA5 reanalysis~\citep{era5}, a global product on a $0.25^\circ$ ($\sim25$\,km) grid, cropped to a bounding box for each region studied. As coarse-grid predictors, we use $20$ dynamic atmospheric variables, including $2$\,m temperature and $10$\,m eastward and northward wind components, and $13$ static surface properties, such as geopotential and other coarse land-surface variables. The full predictor set is listed in \Cref{tab:predictors}. In \cref{sec:aurora}, we replace ERA5 dynamic fields with Aurora forecast grids~\citep{aurora} generated at the same $0.25^\circ$ resolution.

\paragraph*{Station observations.}
Targets are point observations from the Global Historical Climatology Network--Hourly (GHCNh) station network~\citep{ghcnh}. We predict two instantaneous near-surface variables at 6-hourly UTC snapshots: $2$\,m temperature (t2m) and $10$\,m wind speed. Unlike \citet{vaughan2021}, who target daily aggregates, we use instantaneous sub-daily observations because they align directly with forecasting-system outputs at individual lead times and are therefore relevant quantities for the forecast-driven setting of \cref{sec:aurora}.

We evaluate across five climatically diverse regions---Europe, the United States, East Asia, Southern Africa, and Australia---chosen to span a wide range of terrain, land cover, and station density. We use GHCNh rather than the curated VALUE~\citep{gutierrez2019} or ECA\&D station sets, which cover only Europe and were assembled for daily aggregates.

\paragraph*{Topography.}
Following \citet{vaughan2021}, we retain an explicit topographic descriptor, $\mathbf{e}$, comprising three features: elevation; $\Delta\text{elevation}$, defined as the difference between true elevation at the off-grid location and linearly interpolated ERA5 orographic height, obtained from surface geopotential; and the multi-scale topographic position index, $\text{mTPI}$~\citep{mtpi}, which indicates whether a location lies in a valley or on a ridge.

\paragraph*{Hand-crafted surface descriptors.}
As a richer non-learned complement to $\mathbf{e}$, we adopt the station-descriptor set of \citet{bakketun2026}, denoted $\mathbf{s}$, which augments topography with explicit land-cover, vegetation, soil and neighborhood-terrain features (\Cref{tab:extradesc-features}). Its $17$ features divide into two groups. A \emph{surface} group of ten features is aggregated over a $320$\,m radius from ESA WorldCover~v200 \citep{worldcover2021}, ETH global canopy-height product \citep{lang2023canopy}, and SoilGrids at $0$--$5$\,cm depth \citep{soilgrids2021}. A \emph{terrain} group of seven features is aggregated over a $6.25$\,km radius from Copernicus GLO-30 \citep{copernicusdem}, resampled to $250$\,m. The terrain group supplies neighborhood elevation statistics, slope and directional gradients that $\mathbf{e}$ does not contain.

\paragraph*{Land-surface embeddings.}
\tessera{}~\citep{tessera2025} is trained via self-supervised learning to provide a $128$ dimensional embedding for any $10$\,m pixel from a year-long time series of Sentinel-1(2) observations. Each embedding summarizes characteristic annual surface dynamics at a location. To limit storage requirements, we use the 2017 embedding map for all training, validation, and test snapshots, generated using the 1B-parameter model from \tessera{} v2~\citep{tesseraV2}. The downscaler is trained on approximately a decade of historical data and tested on 2022, with the embedding year lying within the training period. This assumes the surface characteristics and seasonal patterns encoded in 2017 remain sufficiently stable across the study period; changes in land cover, vegetation, snow regime, or the built environment are therefore not represented. 

\subsection{Patch compression of \tessera{} embeddings}
\label{sec:method-tessera}

A single \tessera{} pixel describes a $10$\,m spot, which is too local to capture all surface properties affecting a near-surface weather measurement. However, directly supplying all $128$-dimensional embeddings from a $64\times64$ pixel patch introduces a $524{,}288$-dimensional input, which is impractically large and would risk severe overfitting of the downscaler. Instead, we learn a compact, task-agnostic summary of the neighborhood using an unsupervised variational autoencoder (VAE), rather than relying on the downscaler itself to learn this bottleneck from sparse and comparatively limited station observations. Specifically, the VAE compresses a $\sim\,$$640$\,m neighborhood centered on each location into a $16$-dimensional vector $\mathbf{z}_T$ that captures terrain context, land-cover heterogeneity, and coastal--inland transitions.

The patch is deliberately local: it characterizes the immediate surroundings that control how much a weather station's observations depart from the cell mean, rather than summarising the full $\sim\,$$25$\,km cell. The compression rate is severe ($2^{15}$) and deliberately simple, with larger latent representations, multi-scale patches, stronger encoder architectures, and end-to-end VAE fine-tuning left as natural extensions.

\subsection{Downscaling model}
\label{sec:method-model}

The backbone is the convolutional conditional neural process of \citet{gordon2019}, as instantiated for weather downscaling by \citet{vaughan2021}. Its input is a coarse atmospheric grid $Z \in \mathbb{R}^{D_{\mathrm{lat}}\times D_{\mathrm{lon}}\times C}$,
where $D_{\mathrm{lat}}$ and $D_{\mathrm{lon}}$ are the numbers of latitude and longitude grid points in the regional crop, and $C$ is the number of predictor channels at each grid point, as listed in \Cref{tab:predictors}. The model maps $Z$ to a predictive distribution $p(\ystar \mid \mathbf{\xstar}, Z)$ at an arbitrary query location $\mathbf{\xstar}=(\text{lat}, \text{lon})$, where $\ystar$ denotes the corresponding target weather observation, in three stages.

First, $Z$ is passed through a convolutional encoder, producing a feature vector at every grid point, $H = \mathrm{CNN}(Z)$. We write $\mathbf{h}_{nm}$ for the feature vector at the grid point indexed by $n$ along longitude and $m$ along latitude, with $\Delta x_1$ and $\Delta x_2$ the corresponding grid spacings, so that this point sits at coordinates $(n\Delta x_1,\, m\Delta x_2)$. 

Second, these gridded features are evaluated at the off-grid target $\mathbf{\xstar}=(x^{(1)},\, x^{(2)})$ by convolution with a smooth exponentiated-quadratic kernel $\varphi$,
\begin{equation}
  \varphi_c(H,\,\mathbf{\xstar}) \;=\; \sum_{n,m} h_{nm}\,
  \exp\!\left(
    -\frac{\big(\xstar^{(1)} - n\Delta x_1\big)^{2}}{2\ell_1^{2}}
    -\frac{\big(\xstar^{(2)} - m\Delta x_2\big)^{2}}{2\ell_2^{2}}
  \right),
  \label{eq:eq-kernel}
\end{equation}
where $\ell_1,\ell_2$ are learned length-scales. The sum runs over grid points of the regional crop, each weighted by its distance to $\mathbf{\xstar}$, so contributions from nearby cells dominate and the result is defined for any $\mathbf{\xstar}$, on or off the grid, carrying no subgrid information.

Third, an MLP maps this representation, concatenated with per-location descriptors, to the parameters $\boldsymbol{\theta}$ of the predictive distribution,
\begin{equation}
  \boldsymbol{\theta}(\mathbf{\xstar},\,Z,\,\mathbf{d}) \;=\; \psi_{\mathrm{MLP}}\!\big[
  \,\varphi_c(H,\,\mathbf{\xstar})\;,\;\mathbf{d}(\mathbf{\xstar})\,\big],
\label{eq:injection}
\end{equation}
where $\mathbf{d}(\mathbf{\xstar})$ is the \emph{station descriptor}, the concatenation of whichever per-location features a given model variant receives,
\begin{equation}
  \mathbf{d}(\mathbf{\xstar}) \;=\; \big[\,\mathbf{e}(\mathbf{\xstar})\;,\;\mathbf{s}(\mathbf{\xstar})\;,\;\mathbf{z}_{T}(\mathbf{\xstar})\,\big],
\label{eq:descriptor}
\end{equation}
with $\mathbf{e}$ the three-feature topographic descriptor, $\mathbf{s}$ the $17$-feature hand-crafted surface descriptor, and $\mathbf{z}_{T}$ the learned $16$-dimensional \tessera{} surface descriptor. The model variants compared in this paper share the architecture of \cref{eq:injection} entirely and differ only in which features $\mathbf{d}$ contains.  

For brevity, we write $\boldsymbol{\theta}$ hereafter, with its dependence on $\mathbf{\xstar}$, $Z$, and $\mathbf{d}$ left implicit. Since $Z$ contains information only at coarse-grid scale, no mapping of its encoded representation can recover sub-grid structure, leaving $\mathbf{d}$ as the only local, station-specific information available. 

\paragraph*{Predictive heads and training objective.}
Given a context grid $Z$, let $\mathbf{\xstar}^{(1:N)} \equiv \{\mathbf{\xstar}^{(n)}\}_{n=1}^{N}$ denote the set of $N$ target locations and $\ystar^{(1:N)} \equiv \{\ystar^{(n)}\}_{n=1}^{N}$ the corresponding observations.  Following the conditional neural process formulation, we model each target prediction through a separate conditional marginal $p_{\boldsymbol{\theta}}\big(\ystar\, \mid \, \mathbf{\xstar} \,, Z \big)$, where $\boldsymbol{\theta}$ is the parameter vector produced at that location by \cref{eq:injection}. Training minimizes mean negative marginal log likelihood across valid target observations,
\begin{equation}
\ell(\boldsymbol{\theta}\,;\ystar^{(1:N)})
=
- \frac{1}{N}
\sum_{n=1}^{N}
\log
p_{\boldsymbol{\theta}}\big(\ystar^{(n)}\, \mid \, \mathbf{\xstar}^{(n)} \, , Z \big).
\label{eq:marginal-objective}
\end{equation}
The normalization ensures that snapshots contribute equally to training regardless of how many stations report valid observations at that time. Equivalently, this objective corresponds to a factorized conditional joint over target observations, yielding predictive marginals at arbitrary locations but not explicitly modeling residual dependence between targets. In particular, coherent joint uncertainty is not represented: shared synoptic errors can induce correlated residuals across neighboring stations even after conditioning on $Z$. Because the network and decoder parameters are shared across targets, however, nearby locations still receive smoothly varying marginal predictions. All reported evaluation metrics are per-observation marginal metrics.

Both variables use the same two-parameter form $\boldsymbol{\theta}=(\mu,\sigma)$, differing only in distribution family. For $2$\,m temperature, $p_{\boldsymbol{\theta}}$ is a Gaussian $\mathcal{N}(\mu,\sigma^2)$, where per-observation loss is
\begin{equation}
\ell_{\mathrm{Gauss}}(\mu,\sigma;\ystar)
\equiv - \log p_{\boldsymbol{\theta}}\big(\ystar\, \mid \, \mathbf{\xstar} \, , Z \big)
\;=\;
\tfrac{1}{2}\log\!\big(2\pi\sigma^{2}\big)
+\frac{(\ystar-\mu)^{2}}{2\sigma^{2}}.
\label{eq:nll-gauss}
\end{equation}

For $10$\,m wind speed, a Gaussian is not appropriate, as it assigns probability to negative speeds. Instead, we define $p_{\boldsymbol{\theta}}$ as a Truncated Normal distribution over $\ystar\ge 0$:
\begin{equation}
\ell_{\mathrm{TN}}(\mu,\sigma;\ystar)
\equiv
-\log p_{\boldsymbol{\theta}}\big(\ystar\, \mid \, \mathbf{\xstar} \, , Z \big)
\;=\;
\tfrac{1}{2}\log\!\big(2\pi\sigma^{2}\big)
+\frac{(\ystar-\mu)^{2}}{2\sigma^{2}}
+\log\Phi\!\Big(\tfrac{\mu}{\sigma}\Big).
\label{eq:nll-tn}
\end{equation}
where $\Phi$ is the CDF of the standard Gaussian distribution and $\Phi(\mu/\sigma)$ is the truncation normalizer. It is a well-established predictive family for wind speed introduced by \citet{gneiting2006} and developed as the standard EMOS formulation for non-negative quantities by \citet{thorarinsdottir2010}. 

\subsection{Experimental setup}
\label{sec:method-protocol}

\begin{figure}[t]
  \centering
  \includegraphics[width=\textwidth]{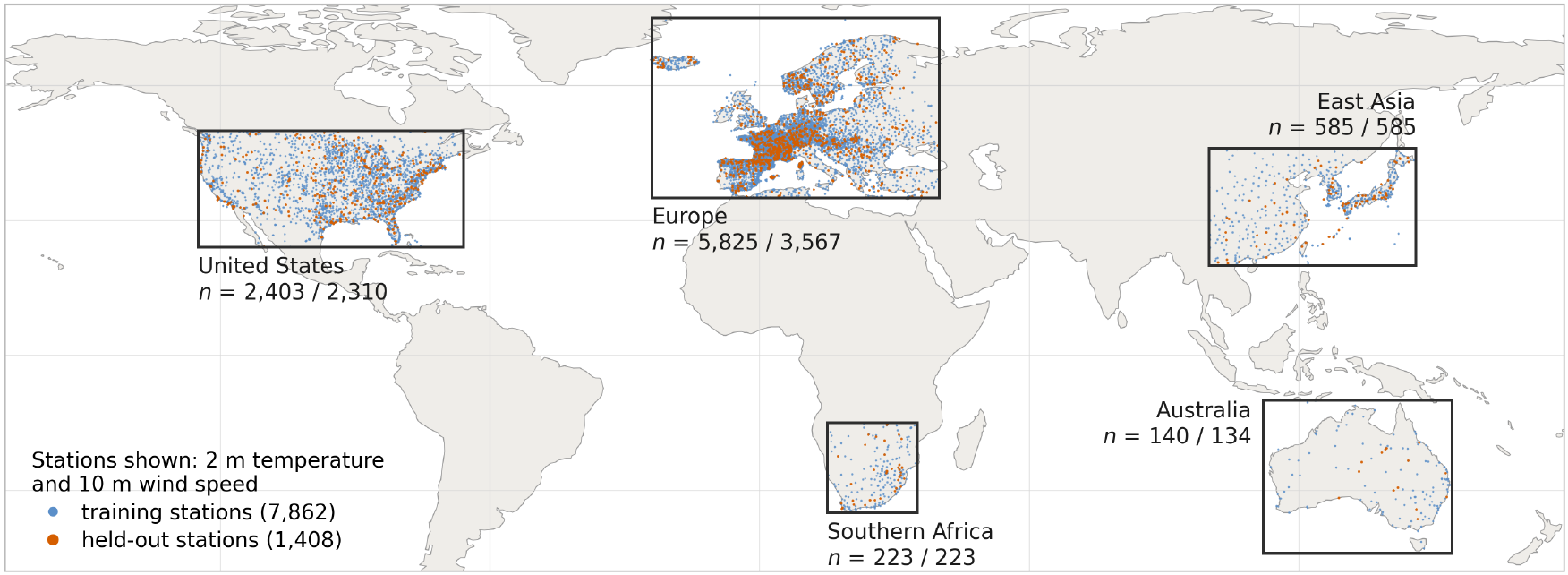}
  \caption{
    The five evaluation regions and their respective station networks. Black rectangles are the ERA5 crop boxes of \Cref{tab:regions}; the United States domain includes some stations in southern Canada, while the Europe domain includes stations in parts of North Africa and western Asia. Dots are the within-region, shortlisted GHCNh stations where: metadata elevation is finite and within $(-900, 8848\rbrack$\,m, rejecting sentinel and physically impossible values; \tessera{} patches can be filled to at least 50\% coverage from the released tiles; and if that patch is not all-zero, nonfinite, or a gross-magnitude outlier. Blue stations enter training, and orange stations are withheld entirely. Region labels give the total station count $n$ as \emph{2\,m~temperature\,/\,10\,m~wind~speed}. Note the more than an order-of-magnitude spread in network density across regions, from Europe to Australia.
  }
  \label{fig:region-overview}
\end{figure}

\paragraph*{Splits.}
We hold out along two axes to evaluate temporal and spatial generalization simultaneously. \emph{Temporally}, 6-hourly snapshots are split by date: $2010$--$2020$ for training, $2021$ for validation, and $2022$ as the test year. \emph{Spatially}, within each region, a fixed random $15\%$ of stations is withheld from training entirely. Every reported score is therefore evaluated at a station the model never saw, on a date it never trained on. \Cref{fig:region-overview} shows the resulting partition, which we reuse across variables and experiments, except in \cref{sec:norway}, where Norwegian stations are withheld on a schedule and occupy a contiguous sub-domain. All results are 3-seeds means.

\paragraph*{Station density.}
To relate a region's skill gain to how sparsely it is observed, we quantify sparsity as station density $\rho$: stations per million square kilometers of the region's bounding box, computed per variable over that region's valid-observation station set. Density spans more than an order of magnitude across our regions (\Cref{fig:region-overview}).

\paragraph*{Model variants.}
The key contribution of this paper is \textit{\convcnp{} with \tessera{}} where $\mathbf{d} = [\mathbf{e},\mathbf{z}_T]$. The ERA5 surface fields are omitted from the context grid $Z$ because $\mathbf{z}_T$ is a richer station-specific surface descriptor, and we found empirically that retaining them degrades held-out performance. We compare it against four references of increasing strength. 

\emph{Persistence} carries forward the station's most recent valid observation, searched backwards in 6-hour steps up to 24\,h, providing a naive lower bound.

\emph{ERA5 interpolation} bilinearly interpolates the coarse field to station locations. For $2$\,m temperature, we additionally correct for elevation difference between a station and interpolated ERA5 orography using $\hat y(\mathbf{\xstar}) \equiv y^{\mathrm{ERA5}}(\mathbf{\xstar}) - \Gamma \cdot \Delta\mathrm{elevation}(\mathbf{\xstar})$~\citep{dutra2020,sheridan2010}, where $\Gamma$ is fitted by least squares on the training split separately for each region. No analogous correction is applied to $10$\,m wind speed since local variations depend on surface properties not generally represented by a fixed correction based only on $\Delta\mathrm{elevation}$~\citep{winstral2017,weng2000}.

The \emph{\convcnp{} (topography-only)} is \citet{vaughan2021}'s original variant and sets $\mathbf{d}=[\mathbf{e}]$. For this baseline, we retain static ERA5 surface fields in the context grid to provide coarse proxies for surface properties otherwise unavailable to the model. 

The \emph{\convcnp{} (hand-crafted surface)} sets $\mathbf{d}=[\mathbf{e},\mathbf{s}]$, adding the $17$-feature descriptor of \cref{sec:method-data}. It receives more topographic and land-surface information than $\mathbf{e}$ alone, testing whether a richer enumeration of physical surface properties can substitute for a learned representation. Similarly to the \tessera{} variant, coarse ERA5 surface fields are omitted. 

\paragraph*{Metrics.}
We report point accuracy with MAE and RMSE, and probabilistic skill with CRPS, evaluated on held-out observations. CRPS is a proper scoring rule that jointly assesses calibration and sharpness, expressed in the target's physical units~\citep{gneiting2007}. We evaluate CRPS in closed form for the Gaussian head, and from $M=200$ draws of the fitted Truncated Normal using the fair estimator~\citep{ferro2008}. Its residual Monte Carlo variance has a negligible effect on aggregated scores. A closed-form solution is also available~\citep{thorarinsdottir2010}, but $\Phi(\mu/\sigma)$ enters the denominator, and we found this leads to loss of numerical precision in low-wind regimes. CRPS is only defined for \convcnp{} variants, since for deterministic references it collapses to MAE.

MAE and RMSE are minimized by the predictive median and mean, which coincide at $\mu$ for the Gaussian head. For the Truncated Normal, the optimal estimators differ and are computed in closed form from $(\mu,\sigma)$, so wind-speed MAE is scored against the median and RMSE against the mean.

\section{Experimental results}
\label{sec:results}
 
This section assesses how \tessera{} affects downscaling performance. We first compare predictive skill across regions and weather variables (\cref{sec:skill}), then examine which descriptor spaces organize subgrid correction (\cref{sec:residual}), and how the embedding changes the fine-scale structure of dense downscaled fields (\cref{sec:maps}). Finally, we test whether these benefits persist when input weather fields are medium-range Aurora forecasts rather than ERA5 (\cref{sec:aurora}), and assess the embedding's sample-efficiency gains during the simulated deployment of a station network in Norway (\cref{sec:norway}).

\subsection{Per-region skill}
\label{sec:skill}

\Cref{tab:main-full} shows that adding \tessera{} embeddings improves over both \convcnp{} baselines in all ten region\,$\times$\,variable settings and across all metrics.

\begin{table}[p]
  \centering
  \caption{Per-region results, ordered by station count. Total station counts $n$ (training $+$ held-out) and network densities $\rho$ given as \emph{t2m\,/\,wind}. Metrics are reported on the held-out stations at held-out years, with the lowest error per region$\times$metric in \textbf{bold}. CRPS is defined only for the probabilistic \convcnp{} heads. Each value is the 3-seeds mean, given the small per-cell cross-seed std ($\le 0.02$ for most entries, rising to ${\sim}0.05$--$0.07$ in Southern Africa and Australia).}
  \label{tab:main-full}
  \footnotesize
  \setlength{\tabcolsep}{4pt}
  \begin{tabular}{l l r r r @{\hspace{1.8em}} r r r}
    \toprule
    & & \multicolumn{3}{c}{\shortstack{\emph{2\,m temperature ($^\circ$C)}}}
      & \multicolumn{3}{c}{\shortstack{\emph{10\,m wind speed (m\,s$^{-1}$)}}} \\
    \cmidrule(lr){3-5}\cmidrule(lr){6-8}
    Region & Model & MAE & RMSE & CRPS & MAE & RMSE & CRPS \\
    \midrule
    \multirow{5}{*}{\shortstack[l]{Europe\\[3pt]{\scriptsize $n=5{,}825/3{,}567$}\\[1pt]{\scriptsize $\rho=322/197$}}}
      & Persistence                & 3.53 & 4.94 & --- & 1.49 & 2.13 & --- \\
      & ERA5 interp. ($+$lapse for t2m)              & 1.35 & 1.95 & --- & 1.44 & 2.15 & --- \\
      & \convcnp{} (topography-only) & 1.17 & 1.67 & 0.84 & 1.28 & 1.87 & 0.92 \\
      & \convcnp{} (hand-crafted surface) & 1.15 & 1.65 & 0.83 & 1.24 & 1.81 & 0.89 \\
      & \convcnp{} with \tessera{} & \textbf{1.10} & \textbf{1.58} & \textbf{0.79} & \textbf{1.19} & \textbf{1.74} & \textbf{0.86} \\
    \addlinespace
    \multirow{5}{*}{\shortstack[l]{United States\\[3pt]{\scriptsize $n=2{,}403/2{,}310$}\\[1pt]{\scriptsize $\rho=159/153$}}}
      & Persistence                & 3.80 & 5.29 & --- & 1.68 & 2.32 & --- \\
      & ERA5 interp. ($+$lapse for t2m)              & 1.52 & 2.23 & --- & 1.63 & 2.12 & --- \\
      & \convcnp{} (topography-only) & 1.41 & 2.07 & 1.03 & 1.45 & 1.96 & 1.04 \\
      & \convcnp{} (hand-crafted surface) & 1.38 & 2.02 & 1.01 & 1.42 & 1.91 & 1.01 \\
      & \convcnp{} with \tessera{} & \textbf{1.30} & \textbf{1.95} & \textbf{0.95} & \textbf{1.36} & \textbf{1.84} & \textbf{0.97} \\
    \addlinespace
    \multirow{5}{*}{\shortstack[l]{East Asia\\[3pt]{\scriptsize $n=585/585$}\\[1pt]{\scriptsize $\rho=47/47$}}}
      & Persistence                & 3.07 & 4.05 & --- & 1.46 & 2.04 & --- \\
      & ERA5 interp. ($+$lapse for t2m)               & 1.22 & 1.70 & --- & 1.44 & 1.99 & --- \\
      & \convcnp{} (topography-only) & 1.35 & 1.87 & 0.97 & 1.23 & 1.71 & 0.88 \\
      & \convcnp{} (hand-crafted surface) & 1.21 & 1.67 & 0.87 & 1.23 & 1.70 & 0.88 \\
      & \convcnp{} with \tessera{} & \textbf{1.10} & \textbf{1.53} & \textbf{0.79} & \textbf{1.16} & \textbf{1.62} & \textbf{0.84} \\
    \addlinespace
    \multirow{5}{*}{\shortstack[l]{Southern Africa\\[3pt]{\scriptsize $n=223/223$}\\[1pt]{\scriptsize $\rho=50/50$}}}
      & Persistence                & 5.36 & 6.90 & --- & 1.48 & 2.07 & --- \\
      & ERA5 interp. ($+$lapse for t2m)               & \textbf{1.69} & 2.56 & --- & 1.38 & 1.83 & --- \\
      & \convcnp{} (topography-only) & 2.00 & 2.75 & 1.45 & 1.23 & 1.65 & 0.88 \\
      & \convcnp{} (hand-crafted surface) & 1.98 & 2.73 & 1.44 & 1.25 & 1.67 & 0.90 \\
      & \convcnp{} with \tessera{} & 1.79 & \textbf{2.55} & \textbf{1.31} & \textbf{1.16} & \textbf{1.57} & \textbf{0.84} \\
    \addlinespace
    \multirow{5}{*}{\shortstack[l]{Australia\\[3pt]{\scriptsize $n=140/134$}\\[1pt]{\scriptsize $\rho=8.9/8.5$}}}
      & Persistence                & 5.60 & 6.62 & --- & 2.03 & 2.61 & --- \\
      & ERA5 interp. ($+$lapse for t2m)               & 1.40 & 2.06 & --- & 1.46 & 1.88 & --- \\
      & \convcnp{} (topography-only) & 1.57 & 2.16 & 1.15 & 1.42 & 1.81 & 1.02 \\
      & \convcnp{} (hand-crafted surface) & 1.50 & 2.16 & 1.12 & 1.33 & 1.71 & 0.96 \\
      & \convcnp{} with \tessera{} & \textbf{1.32} & \textbf{1.92} & \textbf{0.98} & \textbf{1.29} & \textbf{1.69} & \textbf{0.95} \\
    \addlinespace
    \midrule
    \multirow{5}{*}{\emph{All regions}}
      & Persistence                & 4.27 & 5.56 & --- & 1.63 & 2.23 & --- \\
      & ERA5 interp. ($+$lapse for t2m)               & 1.44 & 2.10 & --- & 1.47 & 1.99 & --- \\
      & \convcnp{} (topography-only) & 1.50 & 2.10 & 1.09 & 1.32 & 1.80 & 0.95 \\
      & \convcnp{} (hand-crafted surface) & 1.44 & 2.04 & 1.05 & 1.30 & 1.76 & 0.93 \\
      & \convcnp{} with \tessera{} & \textbf{1.32} & \textbf{1.90} & \textbf{0.96} & \textbf{1.23} & \textbf{1.69} & \textbf{0.89} \\
    \bottomrule
  \end{tabular}
\end{table}

In the densely observed European and US regions, the original \convcnp{} outperforms lapse-corrected interpolation for $2$\,m temperature. However, it is worse in sparser regions, suggesting that topography alone is insufficient to learn a local correction that transfers spatially when training stations provide limited spatial coverage. Adding subgrid surface information reverses this pattern in East Asia and Australia, with \convcnp{} with \tessera{} nearly closing the gap in Southern Africa, remaining worse on MAE only. For $10$\,m wind speed, all three \convcnp{} variants outperform interpolation in every region, with \tessera{} consistently providing a further gain. 

The hand-crafted surface descriptor $\mathbf{s}$ improves on the topography-only baseline in most settings but recovers only a minority of \tessera{}'s benefit: averaged across regions it reduces CRPS by $3.2\%$ for $2$\,m temperature and $2.2\%$ for $10$\,m wind speed, against $11.5\%$ and $6.2\%$ for \tessera{}. It does not reach \convcnp{} with \tessera{} in any of the $30$ region\,${\times}$\,variable\,${\times}$\,metric comparisons. Combining it with \tessera{} helps in two data-rich regions but not elsewhere (\cref{app:extradesc}). 

\Cref{app:controls}'s ablations highlight that \tessera{}'s underlying EO representation drives the performance gains, showing that replacing the VAE compression with simpler summary statistics computed over the $64\times64$ patches recovers $84\%$ of \tessera{}'s CRPS improvement for temperature and effectively all of it for wind speed. Additionally, extra capacity in the \tessera{} variant is not a factor since randomly permuting descriptors $\mathbf{z}_T$ across stations removes much of the benefit.

\begin{figure}[t]
  \centering
  \includegraphics[width=0.85\textwidth]{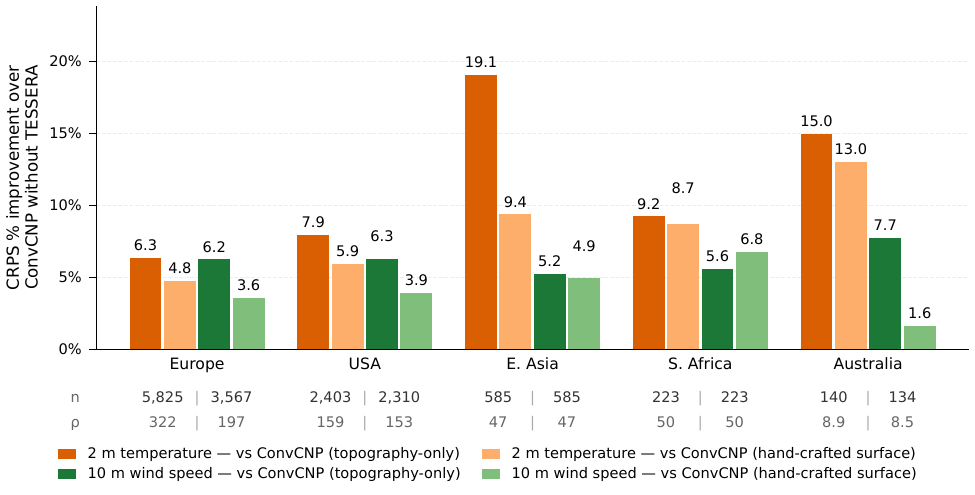}
  \caption{CRPS percentage reduction of \convcnp{} with \tessera{} vs the topography-only and hand-crafted surface baselines, per region, evaluated on the held-out stations at held-out years. Regions are ordered by station count, high to low; $n$ and $\rho$ are the total network size and density for each variable.}
  \label{fig:tessera-crps-uplift}
\end{figure}

\Cref{fig:tessera-crps-uplift} plots percentage improvement of \convcnp{} with \tessera{} over both the topography-only and richer hand-crafted surface baselines. For $2$\,m temperature, the uplift generally increases as station availability decreases against both controls. Together with the comparison against lapse-corrected ERA5 interpolation, this suggests that the embedding acts as a transferable land-surface prior: where observations are spatially abundant, much of the local correction can be learned from station data and hand-crafted surface descriptors, whereas in sparsely observed regions, \tessera{} supplies information that these baselines cannot estimate or represent as effectively.

For $10$\,m wind speed, the uplift over the topography-only baseline is approximately constant across station counts, bar the noisy regions with few held-out stations. The advantage over the hand-crafted surface baseline is less uniform, but persists across regions, showing that richer explicit surface information does not close the gap to \tessera{}. These results suggest that topography provides an incomplete representation for local wind correction, while the learned EO representation makes additional surface information more useful for downscaling. 

\medskip

\subsection{Residual structure: what organizes the subgrid correction}
\label{sec:residual}

The skill differences in \cref{sec:skill} raise the question of which location properties best organize the required correction. We therefore ask, independently of any trained \convcnp{}, which static descriptor space makes the ERA5-interpolation residual predictable at unseen stations. Specifically, we fit the same nonlinear regression probe in geographical, topographic, ERA5-static, hand-crafted surface, and \tessera{} descriptor spaces, and compare their $R^2$ performance. The residual construction, regression procedure, and additional diagnostics are detailed in \cref{app:residual-structure}.

\begin{figure}[t]
  \centering
  \includegraphics[width=0.85\textwidth]{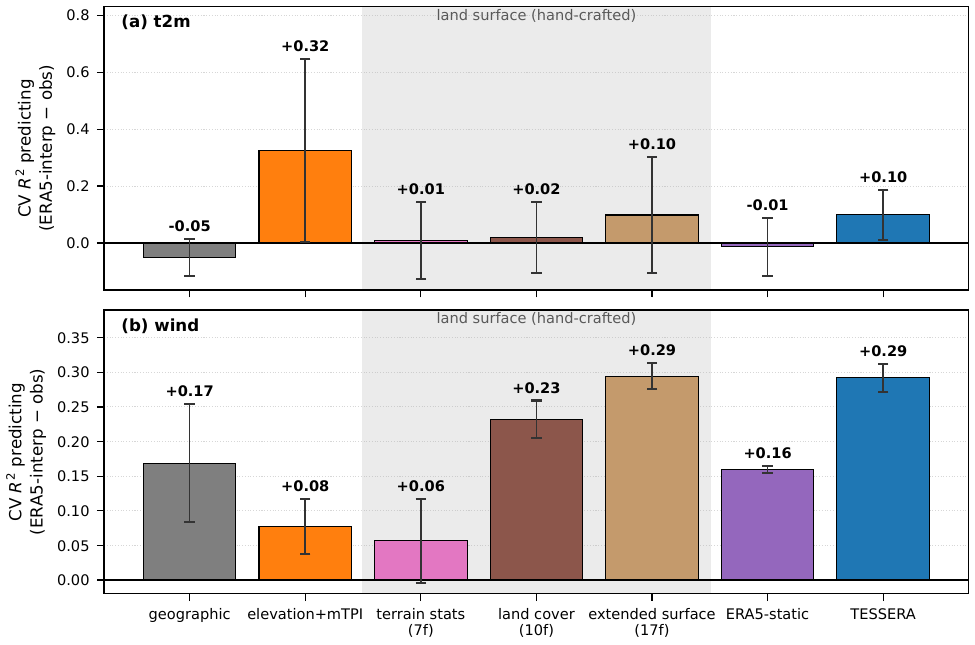}
  \caption{Cross-validated $R^2$ of random-forest regressors fitted independently in each descriptor space on the persistent ERA5-interpolation residual, the mean signed interpolation error at each station over the test snapshots. Bars are the mean over the well-sampled regions (Europe and the United States, $n\ge100$ stations); the residual construction and regression procedure are given in \cref{app:residual-structure}. The three shaded spaces come from the 17-feature hand-crafted descriptor of \citet{bakketun2026} (Table~\ref{tab:extradesc-features}), entered both whole and as its two halves, \emph{terrain stats} and \emph{land cover}. Positive values indicate transferable residual structure at held-out stations; values near or below zero indicate little improvement over predicting the training-fold mean residual. (a) $2$\,m temperature. (b) $10$\,m wind speed.}
  \label{fig:struct}
\end{figure}

\Cref{fig:struct} shows a clear variable-dependent split. For $2$\,m temperature, topography $\mathbf{e}$ is the strongest organizer of persistent residual, reaching a cross-validated $R^2 = 0.32$, against $0.10$ for \tessera{} and the hand-crafted descriptor, and approximately zero for geography and the ERA5-static fields. The sharp separation indicates the persistent temperature residual is better organized by explicit topography than by any land-surface representation: elevation captures most of the dominant temperature correction, while \tessera{} still contains additional, but weaker, transferable information. This aggregate conceals a strong regional contrast, since the topographic signal is concentrated in Europe and is largely absent from the United States network; \cref{app:residual-structure} relates this to the smaller station-to-orography mismatch there.
 
For $10$\,m wind speed, the land-surface spaces dominate. \tessera{} attains the highest cross-validated $R^2$ ($0.29$), outperforming geography ($0.17$), ERA5-static fields ($0.16$) and elevation\,$+$\,mTPI ($0.08$). This indicates that the persistent wind correction is organized by land-surface properties not represented effectively by elevation alone. Notably, the hand-crafted descriptor is not information-poor in this respect: taken as a whole, it matches the embedding exactly ($0.29$).

Read against \Cref{tab:main-full}, \Cref{fig:struct} reinforces the value of a \textit{learned} surface descriptor. The hand-crafted descriptor $\mathbf{s}$ successfully organizes the residuals, but this does not translate into comparable probabilistic skill when used to condition the downscaler: it recovers only about a third of \tessera{}'s CRPS gain. Thus, while $\mathbf{s}$ captures persistent, time-averaged structure in the residuals, \tessera{}'s learned representation is substantially better at converting that surface information into per-snapshot predictive distributions. The remaining sections compare against the topography-only baseline, isolating the contribution of the learned surface representation over explicit topography.

\subsection{Dense maps}
\label{sec:maps}

The probe of \cref{sec:residual} ranks descriptor spaces on a static, station-averaged residual. We now ask how this distinction appears in the spatial structure of the models’ outputs by running both \convcnp{} with \tessera{} and the topography-only baseline as dense $0.05^\circ$ ($\sim$5\,km) fields over Iberia and Norway, revealing how much subgrid structure each resolves below the coarse grid. Both models take terrain elevation from SRTM 1-arc-second DEM~\citep{srtm}, accessed via the Mapzen Terrain Tiles dataset hosted on AWS~\citep{aws-terrain-tiles}, sampled at each $0.05^\circ$ cell. As mTPI is not defined at map scale, it is omitted, leaving elevation and $\Delta$elevation as the sole topographic descriptors.

\begin{figure}[t]
  \centering
  \includegraphics[width=0.92\textwidth]{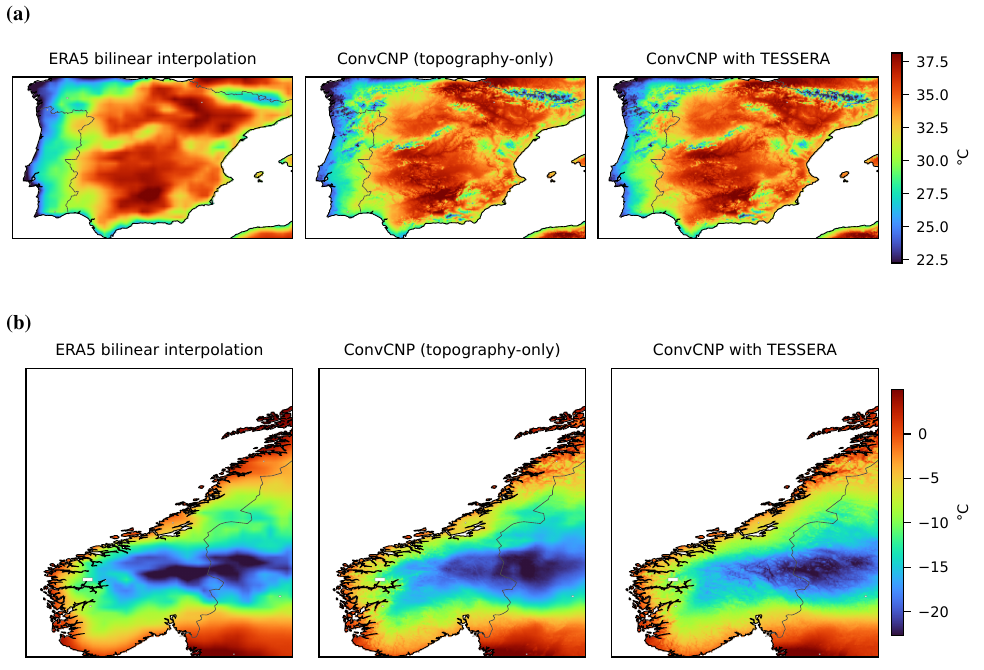}
  \caption{Dense $0.05^\circ$ ($\sim$5\,km) downscaling of 2\,m temperature over (a) Iberia at 1200 UTC 18 July 2022 and (b) Norway at 0000 UTC 2 January 2022. Within each map, columns show (left to right) ERA5 bilinear interpolation, \convcnp{} (topography-only), and \convcnp{} with \tessera{}, each overlaid on a high-resolution DEM. The two \convcnp{} variants resolve similar overall fine-scale amplitude.}
  \label{fig:t2m-maps}
\end{figure}

\begin{figure}[t]
  \centering
  \includegraphics[width=0.92\textwidth]{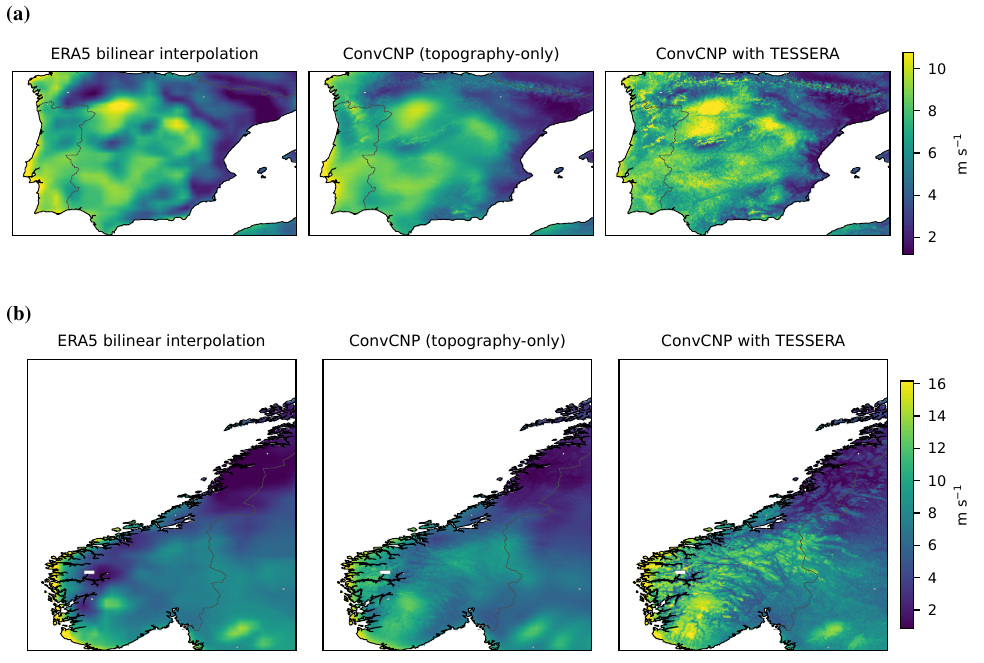}
  \caption{As in Fig.~\ref{fig:t2m-maps}, but for 10\,m wind speed over (a) Iberia at 1200 UTC 12 December 2022 and (b) Norway at 0000 UTC 30 January 2022. The \tessera{} model resolves substantially stronger fine-scale variation, indicating that the embedding exposes spatial information not captured effectively by elevation alone.}
  \label{fig:wind-maps}
\end{figure}

To quantify fine-scale structure visible in Figs.~\ref{fig:t2m-maps} and~\ref{fig:wind-maps}, we measure the amplitude of variation below the coarse-grid scale and compare two model variants (\cref{app:texture-analysis}). 

\begin{figure}[t]
  \centering
  \includegraphics[width=0.92\textwidth]{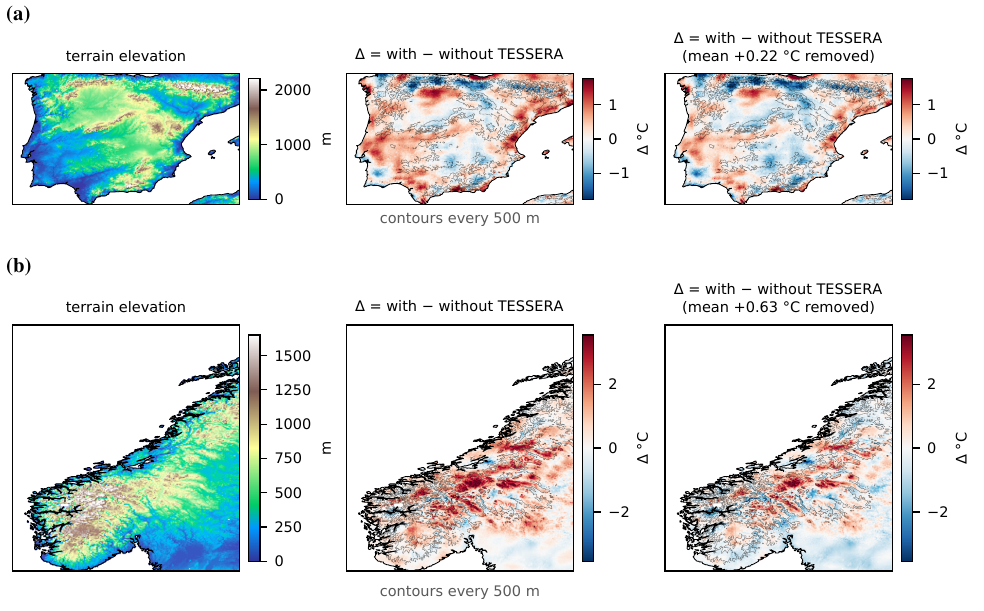}
  \caption{Spatial structure of the change in predictive mean induced by \tessera{} for 2\,m temperature over (a) Iberia and (b) Norway. Within each map, columns show (left to right) terrain elevation, the raw increment, and the mean-removed increment; both increment panels are relief-shaded and carry elevation contours at 500-m intervals. Although the total fine-scale amplitude changes little, the embedding introduces coherent spatial adjustments to the predictive mean.}
  \label{fig:t2m-increments}
\end{figure}

\begin{figure}[t]
  \centering
  \includegraphics[width=0.92\textwidth]{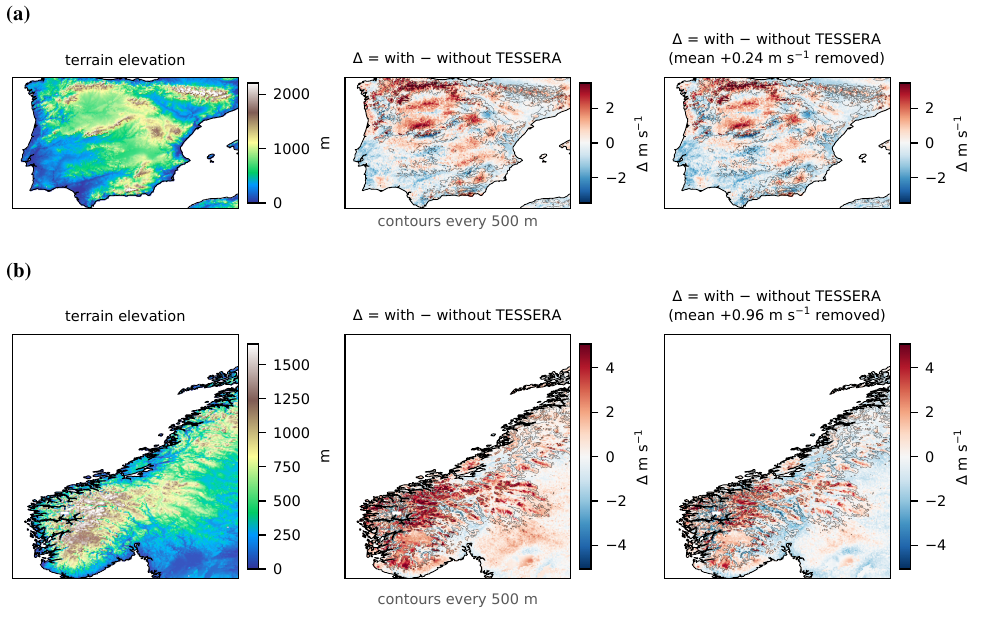}
  \caption{As in Fig.~\ref{fig:t2m-increments}, but for 10\,m wind speed over (a) Iberia and (b) Norway. Strong spatially varying corrections remain in the right-hand panels after the domain-mean increment has been removed.}
  \label{fig:wind-increments}
\end{figure}

\paragraph*{\tessera{} affects temperature and wind differently.}
For $2$\,m temperature, models resolve identical fine-scale amplitudes, with texture ratios of $1.02$ in Iberia and $1.08$ in Norway. Elevation thus captures most of the amplitude of resolved temperature structure. However, increment maps in Figs.~\ref{fig:t2m-increments} and~\ref{fig:wind-increments} reveal coherent local warming and cooling patterns, showing that \tessera{} can modify \textit{where} and by \textit{how much} this structure is expressed without substantially increasing its total amplitude.

For $10$\,m wind speed, \tessera{} increases fine-scale amplitude by $3.28\times$ in Iberia and $3.80\times$ in Norway, with structured increments remaining after removal of the domain-mean shift. Therefore, the embedding contributes location-dependent wind corrections beyond a uniform regional adjustment and beyond the structure represented effectively by elevation alone. Furthermore, independent verification against station networks finds ERA5 near-surface wind to be systematically biased and spatially oversmoothed relative to measurements~\citep{yang2024}, so adding fine-scale amplitude to the wind field is the expected direction of correction. 

These mapped differences are also skillful. Scored at in-region stations on the mapped snapshots (\cref{app:texture-analysis}), \convcnp{} with \tessera{} lowers aggregate CRPS against the topography-only baseline for both variables in both regions (Iberia wind $1.48\!\to\!1.27$\,m\,s$^{-1}$, Norway temperature $1.55\!\to\!1.42\,^{\circ}$C). Additionally, it improves on the baseline at $57$--$65\%$ of individual stations, reflecting a comprehensive gain across the station population rather than large corrections at a few sites. On these snapshots the embedding-induced changes are therefore beneficial on average rather than simply adding unstructured fine-scale variance, though the station counts are small ($n=$~72 wind, 173 temperature) and this should be read alongside the full-test results of \cref{sec:skill}.

The two variables follow \cref{sec:residual}'s conclusions. For $2$\,m temperature, where explicit topography organizes the persistent residual, \tessera{} largely redistributes structure that elevation already places. For $10$\,m wind speed, where the residual is best organized by surface representations, \tessera{} adds fine-scale amplitude, tracking land-cover and surface-roughness boundaries rather than elevation contours. Since both the model-independent probe of time-averaged residuals and the per-snapshot output of a trained downscaler agree on which surface information matters for which variable, the dissociation is likely a property of the near-surface variables rather than an artefact of either analysis.

\subsection{Forecast-driven deployment: from reanalysis to Aurora}
\label{sec:aurora}

An important operational use case for downscaling is producing local predictions from medium-range weather forecasts. Here we replace the ERA5 context with forecasts from the pretrained Aurora~\citep{aurora} checkpoint, a strong deterministic AI weather model. The forecast weather fields are generated at $0.25^\circ$ and initialized from ERA5, storing $+6$, $+24$, and $+72$\,h leads from a single rollout per initialization. Because Aurora does not forecast 6\,h accumulated total precipitation, all input weather grids omit total precipitation at every lead, including lead~0. Rollout storage limits this experiment to two regions: Europe, the largest network, and East Asia, a climatically distinct counterpart with lower station count and network density.

Forecast errors grow and change with lead time, affecting both the expected downscaled value and its uncertainty. A perfect-prognosis downscaler trained only on ERA5 cannot adapt to these lead-dependent forecast errors and may become increasingly biased or under-dispersed at longer leads. We mitigate this by training jointly on all four context grids (ERA5 at lead~0 and the three Aurora leads --- a model output statistics approach), and by passing the normalized horizon (lead$/72$) as an input channel. This allows the downscaler to adjust the full predictive distribution, including both its mean and spread, as a function of lead time.

\begin{figure}[t]
  \centering
  \includegraphics[width=0.85\textwidth]{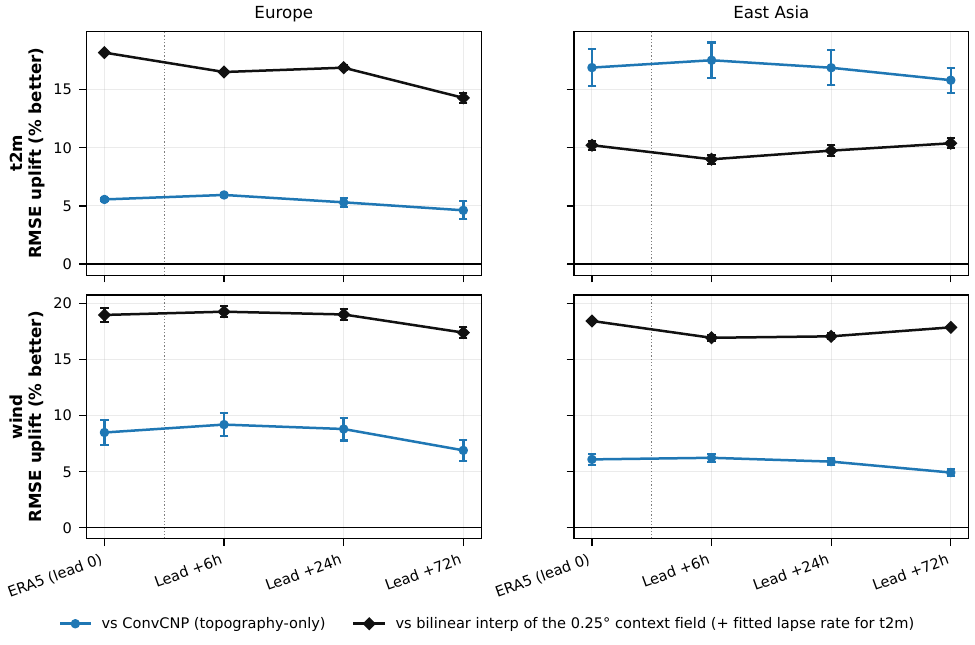}
  \caption{Uplift of \convcnp{} with \tessera{} in RMSE (\% better; positive favours \tessera{}) over two references, by variable (rows) and region (columns). \emph{Blue:} the \convcnp{} (topography-only) baseline. \emph{Black:} bilinear interpolation of the $0.25^\circ$ weather field supplied to the model at that lead --- real ERA5 at lead~$0$, the Aurora forecast at $+6/+24/+72$\,h. For $2$\,m temperature, the interpolation additionally carries the fitted lapse-rate correction $T_{\mathrm{interp}}-\Gamma \cdot \Delta\mathrm{elevation}$ of \cref{sec:skill}. Error bars span the three training seeds.}
  \label{fig:aurora-uplift}
\end{figure}

As per \Cref{fig:aurora-uplift}, \tessera{} provides substantial advantages when downscaling medium-range forecasts, with its benefit persisting across the forecast horizon. Measured in RMSE, it retains lower error than the topography-only baseline at every lead in both regions: for temperature, the uplift is $\sim$6\% in Europe and $\sim$17\% in East Asia at lead~0, and decays only mildly, to $\sim$5\% and $\sim$16\% by $+72$\,h; for wind it is $\sim$6--9\% at lead~0 and eases to $\sim$5--7\% by $+72$\,h. The operational reading is that the land-surface signal the embedding carries is not eroded by forecast error in the weather grid.

\subsection{Sample efficiency under simulated station deployment}
\label{sec:norway}
\tessera{} substantially improves sample efficiency on downstream EO tasks~\citep{tessera2025}. We ask whether it also accelerates extrapolation to previously unobserved sub-regions. Specifically, when a new station network is deployed, how long must it collect observations before its downscaling estimates become reliable, and how well are its locations served \emph{before} observations from the sub-region are available for training? We investigate this through a simulated phased deployment of the Norwegian station network, making time and network-size axes explicit in an operationally realistic scenario.

Norwegian probe stations activate in uniform random order at a constant cadence over a $36$-month window beginning at $T_0 = 2015\text{-}01\text{-}01$, while the non-Norwegian European stations remain available throughout. At each simulated horizon $h$ after $T_0$, we retrain on all observations up to $T_0+h$, each probe contributing only its post-activation record, so the training set $\mathcal{T}(h)$ comprises the non-Norwegian European stations together with probes activated by horizon $h$. Evaluation is fixed throughout on the held-out year $2022$.

\begin{figure}[t]
  \centering
  \includegraphics[width=0.85\textwidth]{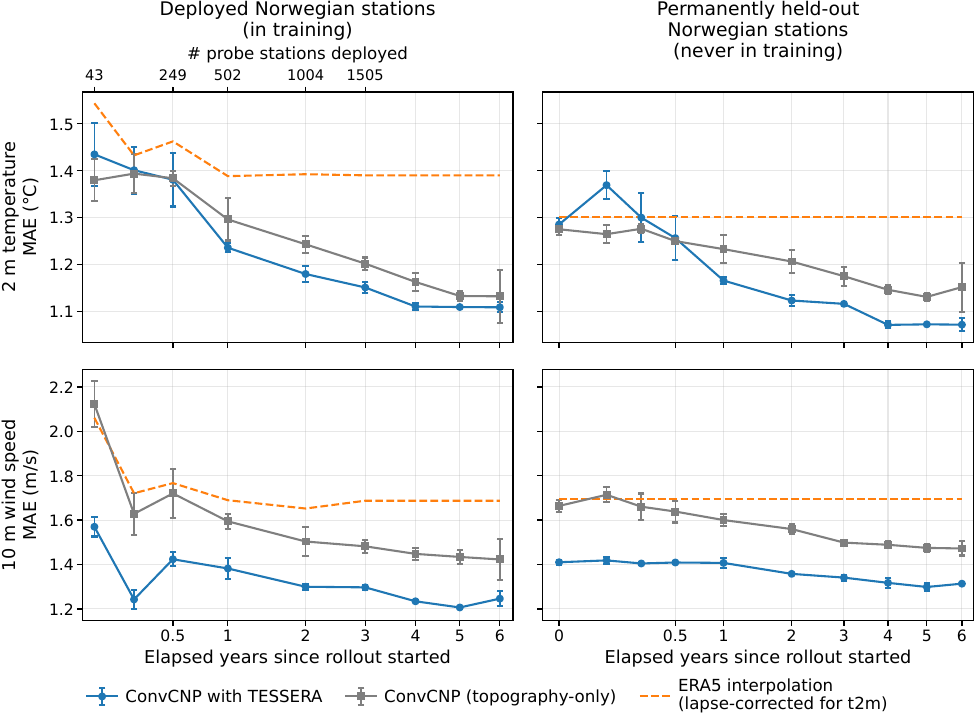}
  \caption{Simulated Norway deployment. MAE on a held-out test year as the Norwegian probe network is progressively deployed into training. Rows: $2$\,m temperature (top), $10$\,m wind speed (bottom). Columns split the Norwegian stations by their relation to the training set: \emph{(left)} probes already deployed stations at that horizon; \emph{(right)} the permanently held-out Norwegian test stations. Curves compare \convcnp{} with \tessera{} (blue) with the topography-only baseline (grey); points are the mean over three seeds and error bars $\pm 1$ standard deviation. The dashed line is ERA5 bilinear interpolation --- lapse-rate corrected for $2$\,m temperature --- aggregated at each horizon over exactly the stations that panel scores. It is therefore flat on the right, whose station set is fixed, and varies on the left only because the deployed population changes as probes come online. Bottom axis: calendar years elapsed since $T_0$, on a square-root scale. Top axis (left column): the number of Norwegian probes deployed into training.}
  \label{fig:norway_rollout}
\end{figure}

Note that the horizon $h$ expands the training window for \emph{every} station. Non-Norwegian stations contribute observations spanning $[\,2010\text{-}01\text{-}01,\;T_0+h\,]$, carrying five years of history at cold start, eight when the deployment completes at $h=3$ years, and eleven at $h=6$ years. Consequently, the curves in \Cref{fig:norway_rollout} reflect all station observations available at each training horizon and do not isolate the effect of adding newly deployed probe stations.

\paragraph*{\tessera{} improves wind speed downscaling from cold start.}
For wind speed, \Cref{fig:norway_rollout} shows that \convcnp{} with \tessera{} holds a clear advantage throughout the deployment across all station partitions. On permanently held-out Norwegian stations, the gap is widest at and immediately after cold start: \tessera{} reduces MAE by $0.25\,\mathrm{m\,s^{-1}}$ before any probe stations are deployed and by $0.30\,\mathrm{m\,s^{-1}}$ after the first month, corresponding to $\sim$$15$--$17\%$ of the topography-only baseline. The advantage subsequently narrows but does not close, remaining $0.16\,\mathrm{m\,s^{-1}}$ ($11\%$) at full deployment. This suggests that increasing station count alone does not fully compensate for the absence of a descriptor exposing surface structure that governs the local wind-speed correction.

Operationally, against ERA5 interpolation, \convcnp{} with \tessera{} is $16.7\%$ better before a single Norwegian probe is deployed ($1.41$ against $1.69\,\mathrm{m\,s^{-1}}$), whereas the topography-only variant merely starts level with it and needs roughly a year of local deployment to open a clear margin. Six years and $1505$ deployed stations later, it reaches $1.47\,\mathrm{m\,s^{-1}}$ --- still above the error \tessera{} attains with no local stations at all. The same ordering holds under RMSE.

\paragraph*{The temperature benefit is smaller and emerges later.}
For $2$\,m temperature, \tessera{} is $0.10$$^\circ$C worse than topography-only after the first month on held-out Norwegian stations. A consistent advantage appears only from year one onwards, reaching~$\sim$$0.08$$^\circ$C ($7\%$) MAE reduction after two years and approximately holding thereafter. This weaker cold-start effect is consistent with \cref{sec:residual}'s conclusion that temperature's downscaling correction is driven primarily by explicit topography. This does not conflict with \cref{sec:skill} since \tessera{} is most valuable when station scarcity also implies poor coverage of the \textit{relevant} land-surface conditions, whereas the available dense European network contains many stations as close elevation analogues to learn the required correction from.

The same conclusions hold under RMSE and CRPS. At early horizons, results for already deployed Norwegian stations in  \Cref{fig:norway_rollout}'s left column rest on $18$ and $12$ stations for temperature and wind, respectively. Their early non-monotonicity may therefore reflect high sampling variability and possible overfitting, both diminishing as more stations are deployed.

To investigate why wind-speed's advantage appears before local observations are available, \cref{app:norway_app} compares held-out Norwegian stations with the evolving training set in geography, topography, ERA5-static, hand-crafted surface, and \tessera{} descriptor spaces. It shows that \tessera{} places almost all held-out Norwegian stations near a training analogue from cold start ($96$\%) while retaining surface information that distinguishes Norway from the broader European domain (AUC $0.96$). This supports the interpretation that the embedding enables corrections to transfer from previously observed surface analogues.

\section{Discussion and conclusion}
\label{sec:discussion}

We show that a learned Earth observation representation can serve as a general-purpose subgrid descriptor for probabilistic weather downscaling. Across five climatically diverse regions, conditioning a \convcnp{}'s decoder on a \tessera{} patch embedding (\cref{sec:method-tessera}, \cref{sec:method-model}) improves both deterministic and probabilistic skill for $2$\,m temperature and $10$\,m wind speed at held-out weather stations, reducing CRPS by $11.5\%$ and $6.2\%$ across regions, respectively, relative to \citet{vaughan2021}'s topography-only ConvCNP baseline. The advantage persists against the stronger baseline with richer hand-crafted surface descriptors adapted from \citet{bakketun2026} (\cref{sec:skill}), and when downscaling medium-range forecasts from the Aurora AI model (\cref{sec:aurora}). 

For $2$\,m temperature, topography best organizes ERA5-interpolation residual (\cref{sec:residual}), and adding \tessera{} produces little additional texture in densely downscaled fields (\cref{sec:maps}). The embedding's principal benefit is providing a broader surface prior when available station observations are insufficient to learn the correction reliably from topographic analogues alone. For $10$\,m wind speed, evidence differs consistently across the residual analysis (\cref{sec:residual}), dense maps (\cref{sec:maps}), and deployment experiment (\cref{sec:norway}), showing the correction is organized most effectively in \tessera{} space. This suggests topography is a less data-efficient representation of the local correction required for near-surface wind speed, as locations of similar elevation can still differ in land cover, exposure, surface roughness, and built environment.

This distinction reinforces the operational value of \tessera{}. In the simulated Norwegian deployment (\cref{sec:norway}), the wind-speed advantage is largest before any Norwegian observations are available and persists after the local network has been fully deployed. The descriptor-space analysis (\cref{app:norway_app}) suggests why: the embedding places almost all unseen Norwegian stations within the support of existing European surface analogues from the outset, while clearly distinguishing them from the remaining training domain. The skill uplift also survives forecast degradation through $+72$\,h (\cref{sec:aurora}), indicating the local surface correction and evolving error of the atmospheric context behave independently over the forecast horizons considered here.

Several limitations bound these conclusions. First, experiments cover only instantaneous $2$\,m temperature and $10$\,m wind speed. The relevance of learned surface descriptors likely differs for precipitation, humidity, radiation, and temporally aggregated quantities, and should be established separately. Models are also trained independently by variable and region, not exploiting physical dependence between variables or testing whether one globally trained downscaler can transfer across climates without regional retraining. Additionally, GHCNh's station network is not distributed randomly over the surface, so performance at uninstrumented environments that are poorly represented even in \tessera{} space remains uncertain.

Second, \tessera{} is treated as a frozen, static surface representation. Each $\mathbf{z}_T$ compresses a fixed $640$\,m patch using a separately trained VAE, and neither the patch scale nor the $16$-dimensional bottleneck is optimized jointly for the downscaling target. This deliberate decoupling limits overfitting, but it does not establish the chosen neighborhood scale or patch encoding objective as optimal. Other EO foundation embeddings may also expose complementary aspects of terrain, land cover, soil properties, and seasonally varying surface state. Time-indexed EO representations (~\cite{olmoearth}) could also capture changes in vegetation, snow cover, water extent, or urban development that a static descriptor cannot represent, provided that genuine time-varying surface information can be separated from leakage or mismatched observation dates.

Third, the model's predictive distribution provides marginal distributions at arbitrary points, but it does not represent residual dependence between nearby stations. This is a property of the predictive construction since the surface descriptor is agnostic to how dependence between targets is modeled. Latent-variable neural processes can recover coherent joint structure at the cost of approximate inference~\citep{convnp}, while a trained \convcnp{} can be deployed autoregressively to produce dependent joint predictions without architectural changes~\citep{bruinsma2023autoregressive}. Alternatively, diffusion-based neural processes offer richer multivariate structure~\citep{dutordoir2023}. Applications requiring joint uncertainty (e.g., aggregate wind-power risk across multiple sites) are therefore a natural extension of the \tessera{}-conditioned downscaler.

This paper's central contribution is therefore not only an improvement to one downscaling architecture, but evidence for a broader role of Earth observation foundation models in weather prediction. To our knowledge, this is the first study to demonstrate that a frozen EO foundation representation can serve as an effective point-specific surface descriptor for probabilistic downscaling at previously unseen locations. In the forecasting setting, it also demonstrates how two independently trained foundation models can be composed: Aurora represents the evolving atmospheric state, while \tessera{} represents the local surface through which that state is expressed. Their combination yields more skilful local predictions than either the coarse forecast or a topography-conditioned downscaler alone, establishing EO embeddings as a practical interface between global weather models and site-specific forecasting.

\section*{Acknowledgments}
Richard E. Turner is supported by the EPSRC Probabilistic AI Hub (EP/Y028783/1), and Will Tebbutt was funded under the Horizon Europe grant 101213369 (DVPS). We gratefully acknowledge unrestricted gifts from John Bernstein and Jane Street. We also thank Melanie Bieli and Albert Wu from Jane Street for their valuable feedback on the manuscript. Pedro Sousa is supported by G-Research.

\datastatement
All datasets analyzed in this study are existing public resources, openly available at the locations given in their citations in the reference list: ERA5~\citep{era5,weatherbench2}, GHCNh station observations~\citep{ghcnh}, the Aurora pretrained checkpoint~\citep{aurora}, and the static surface and terrain products of \cref{sec:method-data}. \tessera{} embeddings~\citep{tesseraV2} are distributed through the GeoTessera library~\citep{geotessera}.

Code for constructing the \tessera{} patch descriptors, training the patch-compression VAE and the \convcnp{} downscaler, and reproducing the figures and tables in this paper is available at \url{https://github.com/pedro15sousa/tessera-downscaling} and archived as version 1.0.0 on Zenodo under the MIT License~\citep{sousa2026tesseracode}. The repository includes the region bounding boxes, station shortlisting criteria, split definitions and random seeds of \cref{sec:method-protocol}.

\amsappendix{app:impl}{Implementation details}

This appendix specifies the coarse predictor set, patch encoder, downscaler architecture, and training configuration. Where not otherwise stated, the ConvCNP construction follows \citet{vaughan2021}.

\subsubsection{Input fields and station descriptors}
\label{app:predictors}

Table~\ref{tab:predictors} lists every channel supplied to the convolutional encoder. Fields are taken from the analysis-ready ERA5 archive distributed with WeatherBench2 \citep{era5,weatherbench2} at 0000, 0600, 1200, and 1800 UTC. Each region is a fixed latitude--longitude box (Table~\ref{tab:regions}), so the encoder sees a regional rather than a global field. The boxes do not overlap, and stations outside all of them are discarded.

\begin{table}[t]
\caption{Coarse predictor channels supplied to the convolutional encoder, on the $0.25^\circ$ ERA5 grid at 6-hourly synoptic times. The choice of upper-air variables and pressure levels follows \citet{vaughan2021}; we differ in using instantaneous rather than daily-maximum 2\,m temperature and in adding 6\,h accumulated precipitation. Static surface fields are the full set of ERA5 time-invariant fields and are supplied to the topography-only baseline only, as per section~2a. Dynamic, static, and coordinate channels are $z$-scored per region using training-split statistics; temporal channels are left unscaled since they are bounded in $[-1,1]$.}
\label{tab:predictors}
\centering
\small
\begin{tabular}{ll}
\hline
Group & Field \\
\hline
\multirow{5}{*}{Surface (dynamic)}
 & 2-m temperature (\texttt{t2m}) \\
 & 10-m eastward wind (\texttt{u10}) \\
 & 10-m northward wind (\texttt{v10}) \\
 & Mean sea level pressure (\texttt{msl}) \\
 & Total precipitation, 6-h accumulation (\texttt{tp}) \\
\hline
\multirow{5}{*}{\shortstack[l]{Upper air (dynamic)\\{\scriptsize 500, 700, 850\,hPa}}}
 & Temperature (\texttt{t}) \\
 & Eastward wind (\texttt{u}) \\
 & Northward wind (\texttt{v}) \\
 & Specific humidity (\texttt{q}) \\
 & Geopotential (\texttt{z}) \\
\hline
\multirow{13}{*}{\shortstack[l]{Static surface\\{\scriptsize baseline only}}}
 & Surface geopotential (\texttt{z}) \\
 & Land--sea mask (\texttt{lsm}) \\
 & Soil type (\texttt{slt}) \\
 & High-vegetation cover (\texttt{cvh}) \\
 & Low-vegetation cover (\texttt{cvl}) \\
 & Type of high vegetation (\texttt{tvh}) \\
 & Type of low vegetation (\texttt{tvl}) \\
 & Angle of subgrid orography (\texttt{anor}) \\
 & Anisotropy of subgrid orography (\texttt{isor}) \\
 & Slope of subgrid orography (\texttt{slor}) \\
 & Std dev of filtered subgrid orography (\texttt{sdfor}) \\
 & Lake cover (\texttt{cl}) \\
 & Lake total depth (\texttt{dl}) \\
\hline
\multirow{2}{*}{Coordinates}
 & Latitude of grid cell \\
 & Longitude of grid cell \\
\hline
\multirow{2}{*}{Temporal}
 & $\cos$, $\sin$ of day of year \\
 & $\cos$, $\sin$ of hour of day \\
\hline
Forecast (section~3d)
 & Normalized lead time (lead$/72$) \\
\hline
\end{tabular}
\end{table}

\begin{table}[t]
\caption{Region bounding boxes. Grid shape is the number of
$0.25^\circ$ ERA5 cells (latitude $\times$ longitude) after cropping.}
\label{tab:regions}
\centering
\small
\begin{tabular}{lccc}
\hline
Region & Latitude & Longitude & Grid shape \\
\hline
Europe          & $35^\circ$--$75^\circ$N   & $24^\circ$W--$40^\circ$E   & $161\times257$ \\
United States   & $24^\circ$--$50^\circ$N   & $125^\circ$--$66^\circ$W   & $105\times237$ \\
East Asia       & $20^\circ$--$46^\circ$N   & $100^\circ$--$146^\circ$E  & $105\times185$ \\
Southern Africa & $35^\circ$--$15^\circ$S   & $15^\circ$--$35^\circ$E    & $81\times81$   \\
Australia       & $44^\circ$--$10^\circ$S   & $112^\circ$--$154^\circ$E  & $137\times169$ \\
\hline
\end{tabular}
\end{table}

Table~\ref{tab:extradesc-features} lists the sources and aggregation radii of the 17 features comprising the hand-crafted surface descriptor $\mathbf{s}$ of \cref{sec:method-data}.

\begin{table}[t]
  \centering
  \caption{The 17 features of the extended hand-crafted descriptor, after \citet{bakketun2026}. Rules in the right-hand columns separate features drawn from different source products; the two feature groups differ in the radius over which they are aggregated.}
  \label{tab:extradesc-features}
  \footnotesize
  \begin{tabular}{l l l l}
    \toprule
    Group & Feature & Units & Source \\
    \midrule
    \multirow{9}{*}{\shortstack[l]{Surface\\{\scriptsize $320$\,m radius}}}
      & \texttt{forest\_frac}   & fraction & \multirow{7}{*}{ESA WorldCover v200~\citep{worldcover2021}} \\
      & \texttt{lowveg\_frac}   & fraction & \\
      & \texttt{crop\_frac}     & fraction & \\
      & \texttt{built\_frac}    & fraction & \\
      & \texttt{bare\_frac}     & fraction & \\
      & \texttt{snowice\_frac}  & fraction & \\
      & \texttt{water\_frac}    & fraction & \\
    \cmidrule(l){2-4}
      & \texttt{tree\_height}   & m        & ETH canopy height, 2020~\citep{lang2023canopy} \\
    \cmidrule(l){2-4}
      & \texttt{clay\_frac}     & g\,kg$^{-1}$ & \multirow{2}{*}{SoilGrids $0$--$5$\,cm~\citep{soilgrids2021}} \\
      & \texttt{sand\_frac}     & g\,kg$^{-1}$ & \\
    \midrule
    \multirow{7}{*}{\shortstack[l]{Terrain\\{\scriptsize $6.25$\,km radius}}}
      & \texttt{elev\_mean}     & m & \multirow{7}{*}{Copernicus GLO-30 at $250$\,m~\citep{copernicusdem}} \\
      & \texttt{elev\_std}      & m & \\
      & \texttt{elev\_min}      & m & \\
      & \texttt{elev\_max}      & m & \\
      & \texttt{slope}          & $\tan(\text{slope})$ & \\
      & \texttt{dz\_dn}         & m\,m$^{-1}$ & \\
      & \texttt{dz\_de}         & m\,m$^{-1}$ & \\
    \bottomrule
  \end{tabular}
\end{table}

\subsubsection{Patch compression VAE}
\label{app:vae}

Per-location \tessera{} surface descriptors $\mathbf{z}_T$ are produced by a convolutional variational autoencoder \citep{kingma2014vae}, trained on station-centered patches and thereafter frozen. Its settings are given in Table~\ref{tab:hyperparams}. At inference, the posterior mean is taken as the descriptor. The encoder sees neither weather observations nor split labels.

\subsubsection{Downscaler architecture and training}
\label{app:convcnp}

Architecture and optimization settings are given in Table~\ref{tab:hyperparams}. One episode is a single snapshot for a single region, with every valid station in that region as a target, and stations with missing observations masked out of the loss.

\begin{table}[t]
\caption{Architecture and optimization settings for the patch-compression VAE and the \convcnp{} downscaler. The gradient loss $\mathcal{L}_\nabla$, evaluated on first-order spatial finite differences of the patch, discourages the decoder from settling on a spatially flat reconstruction matching only the patch mean. The small terminal KL weight was necessary because at weights of order one the approximate posterior converged toward the prior and the latent ceased to carry station-specific information \citep{ichikawa2024}. The kernel length scale is initialized toward locality rather than region-wide smoothing, leaving the optimizer free to widen it if the data support it.}
\label{tab:hyperparams}
\centering
\small
\begin{tabular}{lll}
\hline
Component & Setting & Value \\
\hline
\multirow{12}{*}{\shortstack[l]{Patch-compression\\VAE}}
 & Encoder blocks   & $4\times$ (stride-2 $3\times3$ conv, BatchNorm, ReLU, 2-D dropout 0.1) \\
 & Encoder widths   & $128 \to 256 \to 256 \to 512$ \\
 & Spatial reduction & $64\times64 \to 4\times4$ \\
 & Latent           & 16-dim diagonal Gaussian from flattened $512\times4\times4$ \\
 & Decoder          & Linear to $512\times4\times4$; $512 \to 256 \to 256 \to 128$ \\
 & Decoder output   & Transposed conv to 128 channels, no norm or activation \\
 & Gradient weight $\lambda_\nabla$ & $0.5$ \\
 & KL weight $\beta$ & $0 \to 5\times10^{-4}$ linearly over 5000 steps, then constant \\
 & Log-variance clamp & $[-10, 10]$ \\
 & Optimizer        & AdamW, learning rate $10^{-3}$, weight decay $10^{-5}$ \\
 & Batch size, epochs & 64, 200 (cosine schedule; best validation checkpoint) \\
 & Gradient clipping & Global norm 1.0 \\
\hline
\multirow{4}{*}{\shortstack[l]{\convcnp{}\\encoder}}
 & Stem             & $3\times3$ convolution to 128 features \\
 & Body             & 3 residual blocks, each $2\times$ ($3\times3$ conv, 128 channels, ReLU) \\
 & Head             & $1\times1$ convolution to 128 output channels \\
 & Resolution       & Preserved: no striding, no pooling, zero padding \\
\hline
\multirow{4}{*}{\shortstack[l]{\convcnp{}\\decoder}}
 & Kernel length scale & Learned as $\log\ell$, initialized $\ell = 0.5^\circ$ (${\approx}55$\,km) \\
 & MLP              & $3\times$ (Linear, ReLU), width 128 \\
 & Input width      & 147 with \tessera{} ($128+3+16$); 131 for the baseline \\
 & Trainable parameters & $9.84\times10^5$ with \tessera{}; $9.97\times10^5$ baseline \\
\hline
\multirow{5}{*}{\shortstack[l]{\convcnp{}\\optimization}}
 & Optimizer        & Adam, learning rate $2.5\times10^{-5}$, weight decay $10^{-4}$ \\
 & Warmup           & Linear from $10^{-3}$ of target over first 5\% of steps, then constant \\
 & Gradient handling & Clipped to global norm 1.0; nonfinite steps skipped \\
 & Stopping         & ${\le}100$ epochs, early stopping on validation NLL, patience 10 \\
 & Seeds, hardware  & 42, 123, 456; one NVIDIA GH200 per run (${\approx}9600$ GPU-hours total) \\
\hline
\end{tabular}
\end{table}

\amsappendix{app:extradesc}{Region-level results for the hand-crafted surface descriptor}

\Cref{tab:extradesc} adds a fourth variant, compared to \Cref{tab:main-full}, where the hand-crafted surface descriptor $\mathbf{s}$ is supplied \emph{on top of} \tessera{}, $\mathbf{d}=[\mathbf{e},\mathbf{s},\mathbf{z}_T]$. Splits, seeds, station sets, architecture and optimization are identical across all four variants.

\begin{table}[t]
  \centering
  \caption{Effect of the hand-crafted surface descriptor $\mathbf{s}$, evaluated on the same held-out stations and years as \Cref{tab:main-full}, at three decimal places. Lowest error per region$\times$metric in \textbf{bold}. Values are means over 3 seeds.}
  \label{tab:extradesc}
  \footnotesize
  \setlength{\tabcolsep}{4pt}
  \begin{tabular}{l l r r r @{\hspace{1.5em}} r r r}
    \toprule
    & & \multicolumn{3}{c}{\emph{2\,m temperature ( $^\circ$ C)}}
      & \multicolumn{3}{c}{\emph{10\,m wind speed (m\,s$^{-1}$)}} \\
    \cmidrule(lr){3-5}\cmidrule(lr){6-8}
    Region & Model & MAE & RMSE & CRPS & MAE & RMSE & CRPS \\
    \midrule
    \multirow{4}{*}{Europe}
      & \convcnp{} (topography-only)         & 1.169 & 1.673 & 0.843 & 1.283 & 1.874 & 0.918 \\
      & \quad $+$ hand-crafted surface      & 1.152 & 1.647 & 0.829 & 1.244 & 1.813 & 0.893 \\
      & \convcnp{} with \tessera{}         & 1.098 & 1.577 & 0.789 & 1.191 & 1.739 & 0.861 \\
      & \quad $+$ hand-crafted surface      & \textbf{1.066} & \textbf{1.522} & \textbf{0.766} & \textbf{1.182} & \textbf{1.709} & \textbf{0.845} \\
    \addlinespace
    \multirow{4}{*}{United States}
      & \convcnp{} (topography-only)         & 1.413 & 2.070 & 1.029 & 1.446 & 1.960 & 1.038 \\
      & \quad $+$ hand-crafted surface      & 1.381 & 2.023 & 1.007 & 1.416 & 1.910 & 1.013 \\
      & \convcnp{} with \tessera{}         & 1.299 & 1.946 & 0.947 & 1.362 & 1.841 & 0.973 \\
      & \quad $+$ hand-crafted surface      & \textbf{1.280} & \textbf{1.918} & \textbf{0.935} & \textbf{1.353} & \textbf{1.818} & \textbf{0.965} \\
    \addlinespace
    \multirow{4}{*}{East Asia}
      & \convcnp{} (topography-only)         & 1.347 & 1.867 & 0.974 & 1.231 & 1.705 & 0.882 \\
      & \quad $+$ hand-crafted surface      & 1.207 & 1.669 & 0.870 & 1.233 & 1.697 & 0.879 \\
      & \convcnp{} with \tessera{}         & \textbf{1.097} & \textbf{1.530} & \textbf{0.788} & 1.164 & 1.622 & 0.836 \\
      & \quad $+$ hand-crafted surface      & 1.101 & \textbf{1.530} & 0.790 & \textbf{1.149} & \textbf{1.611} & \textbf{0.825} \\
    \addlinespace
    \multirow{4}{*}{Southern Africa}
      & \convcnp{} (topography-only)         & 1.998 & 2.749 & 1.447 & 1.231 & 1.646 & 0.885 \\
      & \quad $+$ hand-crafted surface      & 1.980 & 2.725 & 1.438 & 1.250 & 1.671 & 0.896 \\
      & \convcnp{} with \tessera{}         & 1.791 & 2.546 & 1.313 & \textbf{1.164} & \textbf{1.566} & \textbf{0.835} \\
      & \quad $+$ hand-crafted surface      & \textbf{1.776} & \textbf{2.540} & \textbf{1.302} & 1.173 & 1.570 & 0.842 \\
    \addlinespace
    \multirow{4}{*}{Australia}
      & \convcnp{} (topography-only)         & 1.573 & 2.164 & 1.149 & 1.422 & 1.812 & 1.025 \\
      & \quad $+$ hand-crafted surface      & 1.498 & 2.159 & 1.123 & 1.331 & 1.713 & 0.961 \\
      & \convcnp{} with \tessera{}         & \textbf{1.323} & \textbf{1.922} & \textbf{0.977} & 1.292 & 1.690 & 0.945 \\
      & \quad $+$ hand-crafted surface      & 1.369 & 1.959 & 1.004 & \textbf{1.279} & \textbf{1.660} & \textbf{0.922} \\
    \addlinespace
    \midrule
    \multirow{4}{*}{\emph{All regions}}
      & \convcnp{} (topography-only)         & 1.500 & 2.105 & 1.088 & 1.323 & 1.799 & 0.949 \\
      & \quad $+$ hand-crafted surface      & 1.444 & 2.044 & 1.053 & 1.295 & 1.761 & 0.928 \\
      & \convcnp{} with \tessera{}         & 1.321 & 1.904 & 0.963 & 1.234 & 1.692 & 0.890 \\
      & \quad $+$ hand-crafted surface      & \textbf{1.319} & \textbf{1.894} & \textbf{0.959} & \textbf{1.227} & \textbf{1.674} & \textbf{0.880} \\
    \bottomrule
  \end{tabular}
\end{table}

Supplying both $\mathbf{s}$ and $\mathbf{z}_T$ helps in the two data-rich regions but is worse in $8$ of $30$ comparisons, concentrated in East Asia, Southern Africa and Australia. Aggregated over all regions, it is essentially indistinguishable from \tessera{} alone for temperature ($1.319$ against $1.321$ MAE) and marginally better for wind ($1.227$ against $1.234$). We therefore do not claim the two descriptors are complementary in general: where stations are plentiful, additional features carry some independent signal, and where they are scarce, \tessera{}'s generalization is preferred.

\amsappendix{app:controls}{Shuffled-descriptor control and simpler patch summaries}

The gain reported in \cref{sec:skill} could in principle arise from adding a $16$-dimensional input --- extra parameters in the decoder's first layer, or a regularizing effect of a high-entropy input --- rather than from the surface information $\mathbf{z}_T$ carries about each station. Second, assuming it carries real signal, the pre-trained VAE encoder of \cref{sec:method-tessera} is only one way to compress the patch, and a far simpler summary might suffice. We address both questions with dedicated controls.

\begin{table}[t]
\centering
\caption{Descriptor controls, evaluated on the same held-out stations and years as \Cref{tab:main-full}, in its identical configuration --- the \emph{none} and \emph{VAE latent} rows reproduce that table's \convcnp{} rows. Within each region, the four \convcnp{} models share architecture, parameter count, splits, and seeds; the three descriptor rows differ only in the content of the 16-dimensional per-station vector: the VAE-learned \tessera{} surface descriptor, the same descriptors under a fixed random station permutation, and summary statistics computed over the same patch of embeddings. Values are means over 3 seeds; Lowest error per region\,$\times$\,metric in \textbf{bold}.}
\label{tab:descriptor-controls}
\small
\begin{tabular}{llcccccc}
\toprule
& & \multicolumn{3}{c}{2\,m temperature ($^\circ$C)} & \multicolumn{3}{c}{10\,m wind speed (m\,s$^{-1}$)} \\
\cmidrule(lr){3-5}\cmidrule(lr){6-8}
Region & Additional surface descriptor & MAE & RMSE & CRPS & MAE & RMSE & CRPS \\
\midrule
\multirow{4}{*}{Europe}
& none (topography-only)     & 1.169 & 1.673 & 0.843 & 1.283 & 1.874 & 0.918 \\
& VAE latent, shuffled     & 1.128 & 1.616 & 0.811 & 1.236 & 1.796 & 0.892 \\
& patch summary statistics & 1.102 & 1.583 & 0.792 & 1.212 & 1.767 & 0.867 \\
& VAE latent               & \textbf{1.098} & \textbf{1.577} & \textbf{0.789} & \textbf{1.191} & \textbf{1.739} & \textbf{0.861} \\
\midrule
\multirow{4}{*}{United States}
& none (topography-only)     & 1.413 & 2.070 & 1.029 & 1.446 & 1.960 & 1.038 \\
& VAE latent, shuffled     & 1.339 & 2.000 & 0.976 & 1.406 & 1.905 & 1.011 \\
& patch summary statistics & 1.315 & 1.975 & 0.960 & 1.365 & 1.847 & 0.977 \\
& VAE latent               & \textbf{1.299} & \textbf{1.946} & \textbf{0.947} & \textbf{1.362} & \textbf{1.841} & \textbf{0.973} \\
\midrule
\multirow{4}{*}{East Asia}
& none (topography-only)     & 1.347 & 1.867 & 0.974 & 1.231 & 1.705 & 0.882 \\
& VAE latent, shuffled     & 1.124 & 1.563 & 0.808 & 1.173 & 1.628 & 0.840 \\
& patch summary statistics & \textbf{1.097} & 1.533 & \textbf{0.788} & \textbf{1.164} & \textbf{1.620} & \textbf{0.833} \\
& VAE latent               & \textbf{1.097} & \textbf{1.530} & \textbf{0.788} & \textbf{1.164} & 1.622 & 0.836 \\
\midrule
\multirow{4}{*}{Southern Africa}
& none (topography-only)     & 1.998 & 2.749 & 1.447 & 1.231 & 1.646 & 0.885 \\
& VAE latent, shuffled     & 1.879 & 2.637 & 1.367 & 1.239 & 1.659 & 0.889 \\
& patch summary statistics & 1.928 & 2.664 & 1.394 & 1.175 & 1.588 & 0.845 \\
& VAE latent               & \textbf{1.791} & \textbf{2.546} & \textbf{1.313} & \textbf{1.164} & \textbf{1.566} & \textbf{0.835} \\
\midrule
\multirow{4}{*}{Australia}
& none (topography-only)     & 1.573 & 2.164 & 1.149 & 1.422 & 1.812 & 1.025 \\
& VAE latent, shuffled     & 1.358 & 1.980 & 1.006 & 1.327 & 1.727 & 0.964 \\
& patch summary statistics & \textbf{1.313} & 1.933 & 0.982 & \textbf{1.265} & \textbf{1.639} & \textbf{0.906} \\
& VAE latent               & 1.323 & \textbf{1.922} & \textbf{0.977} & 1.292 & 1.690 & 0.945 \\
\midrule
\multirow{4}{*}{\emph{All regions}}
& none (topography-only)     & 1.500 & 2.105 & 1.088 & 1.323 & 1.799 & 0.949 \\
& VAE latent, shuffled     & 1.365 & 1.959 & 0.994 & 1.276 & 1.743 & 0.919 \\
& patch summary statistics & 1.351 & 1.937 & 0.984 & 1.236 & \textbf{1.692} & \textbf{0.886} \\
& VAE latent               & \textbf{1.321} & \textbf{1.904} & \textbf{0.963} & \textbf{1.234} & \textbf{1.692} & 0.890 \\
\bottomrule
\end{tabular}
\end{table}

\paragraph*{Shuffled-descriptor control.}
Reassigning each station's $\mathbf{z}_T$ descriptor to another station via a random permutation makes per-station content uninformative. The architecture, parameter count, optimization, and splits are all unchanged, so any benefit surviving shuffling cannot derive from station-specific surface information.

\Cref{tab:descriptor-controls} shows that shuffling degrades CRPS in all ten region\,$\times$\,variable settings, and, averaging over regions, gives back a quarter of the CRPS gap to the topography-only baseline for temperature and half of it for wind speed. We interpret the paired live-versus-shuffled contrast as the controlled estimate of the station-specific contribution. Moreover, because weather stations occupy a non-random subset of surface environments, shuffled surfaces are still drawn from the empirical distribution of station surfaces rather than arbitrary noise. They may therefore retain a broad station-surface prior even after their location-specific alignment is removed.

\paragraph*{Summary statistics instead of the learned patch encoder.}
The second control replaces the VAE with simpler summary statistics of the same $64\times64$-pixel ($\sim$640\,m) patch of \tessera{} embeddings. We select four embedding channels whose per-station spatial means vary most across training stations, and describe each by its mean, standard deviation, and the 10th and 90th percentiles over the patch. This produces a $16$-dimensional vector, matched in dimensionality to the VAE's compressed latent. 

The summary statistics recover most of the gain. Averaged over regions, they retain $84\%$ of the CRPS improvement for temperature and effectively all of it for wind speed, outperforming the topography-only baseline in every region. The learned VAE encoder retains a small but fairly consistent edge, with lower CRPS in seven of ten region\,$\times$\,variable settings. We therefore retain it as the canonical descriptor, although the choice of patch-compression method is evidently not a bottleneck.

\amsappendix{app:residual-structure}{Model-independent analysis of residual structure}

This appendix provides the construction and validation details for the descriptor-space analysis summarized in \cref{sec:residual}. 

\paragraph*{Residual target.}
We predict residuals rather than station values directly. Conceptually, a station observation can be written as
\begin{equation}
    y_{s,t} = y^{\mathrm{interp}}_{s,t} - r_s + \varepsilon_{s,t},
\end{equation}
where $y^{\mathrm{interp}}_{s,t}$ represents the resolved, time-varying atmospheric state, $r_s$ is a persistent location-specific interpolation bias, and $\varepsilon_{s,t}$ collects transient local effects and observation noise. Since candidate descriptors are static properties of station location, asking them to predict $y_{s,t}$ directly would primarily test their ability to explain station climatology while confounding the analysis with the much larger temporal variability of the weather. Subtracting the station observation from ERA5 interpolation removes much of this resolved temporal variability, while averaging over snapshots suppresses transient noise. The resulting target therefore isolates the persistent spatial component that a surface descriptor should organize.

We define the ERA5-interpolation residual at station $s$ as the mean signed interpolation error over its set $T_s$ of valid test snapshots:
\begin{equation}
  r_s
  =
  \frac{1}{|T_s|}
  \sum_{t\in T_s}
  \left(
      y^{\mathrm{interp}}_{s,t}-y_{s,t}
  \right),
  \label{eq:era5-int-residual}
\end{equation}
where $y^{\mathrm{interp}}_{s,t}$ is the bilinearly interpolated ERA5 value and $y_{s,t}$ is the station observation. The negative residual, $-r_s$, is the mean additive correction that a downscaler would need to apply to ERA5 at station $s$. Since reversing the target sign does not affect the relative predictive performance of the regressors, analyzing $r_s$ is equivalent to analyzing the required correction for the purpose of comparing descriptor spaces.

Unlike the residual of a trained \convcnp{}, $r_s$ is constructed without using any station descriptor, making direct comparisons across spaces possible by evaluating against the same target.\footnote{For the ERA5-static descriptor space, we use static fields linearly interpolated to each station, not the model's CNN-encoded grid $Z$. The latter is dominated by time-dependent dynamic weather processed by the same encoder, which is the field being \emph{corrected} at the station, not a persistent driver of local correction. Including it would flag Norway as out-of-distribution for reasons orthogonal to whether a surface analogue exists. The static interpolant isolates persistent surface character.}. 

\paragraph*{Descriptor-space regression probe.}
For each candidate descriptor space $\mathcal{D}$, let
$d_s^{\mathcal{D}}\in\mathbb{R}^{p_{\mathcal{D}}}$ denote its $p_{\mathcal{D}}$-dimensional feature vector at station $s$. We fit a separate random-forest regressor 
\begin{equation}
    \widehat{r}^{\mathcal{D}}_s
    =
    f_{\mathcal{D}}\!\left(d_s^{\mathcal{D}}\right)
\end{equation}
for each descriptor space. The random forest is not intended as the final downscaling model, but as a common nonlinear probe. It tests whether information and geometry of descriptor space $\mathcal{D}$ make the persistent ERA5-interpolation residual predictable at stations excluded from fitting.

Predictive performance is measured using $5$-fold cross-validation and we report the mean of the per-fold $R^2_{\mathrm{CV}}(\mathcal{D})$, ranking which representation most effectively organizes the persistent subgrid residual in the well-sampled regions (Europe and United States).

\paragraph*{Temperature-network diagnostics.}
In the United States, no descriptor attains materially positive performance for temperature. This is consistent with the station network containing substantially less unresolved topographic variation than the European network: $24\%$ of European stations are displaced by more than $200$\,m from linearly interpolated ERA5 orography, and the residual correlates with that displacement at $\rho=0.83$, compared with $5\%$ and $\rho=0.32$ in the United States. A topographic descriptor can organize the interpolation residual only to the extent that unresolved topographic mismatch is present in the evaluated station network.

\amsappendix{app:texture-analysis}{Quantifying and validating fine-scale map structure}

This appendix provides the quantitative analysis underlying the dense-map interpretation in \cref{sec:maps}, measuring how much fine-scale spatial variation each model resolves.

\paragraph*{Texture statistic.}
We measure the structure each field carries below the coarse grid scale. Throughout, $\hat y(\mathbf{x})$ is the predictive marginal mean at cell $\mathbf{x}$ (median for wind), averaged over the three training seeds. The mapped predictive $0.05^\circ$ field is \textbf{not} a draw from the joint predictive $p\big(\ystar^{(1:N)} \mid \xstar^{(1:N)}, Z\big)$.

Writing $\langle\hat{y}\rangle_{\sigma}(x)$ for a Gaussian low-pass of $\hat y$ with standard deviation $\sigma = 3$ map cells ($0.15^\circ$), renormalized over valid cells so that the sea mask does not corrupt the average near coastlines, the fine-scale component is the high-pass residual
\begin{equation}
  r(x) \;=\; \hat{y}(x) \;-\; \langle\hat{y}\rangle_{\sigma}(x),
  \label{eq:highpass}
\end{equation}
and the field's texture is its spatial standard deviation over the $|X|$ mapped cells,
\begin{equation}
  \mathcal{T} \;=\; \sqrt{\tfrac{1}{|X|}\sum_{x\in X}\big(r(x)-\bar r\big)^{2}},
  \qquad \bar r=\tfrac{1}{|X|}\sum_{x\in X} r(x),
  \label{eq:texture}
\end{equation}
i.e.\ the amplitude of variation below the $0.15^\circ$ smoothing scale that the downscaler resolves. We report the ratio $\mathcal{T}_{\tessera}/\mathcal{T}_{\text{base}}$ of the two models' textures.

\amsappendix{app:norway_app}{Descriptor-space coverage of Norwegian probe stations}

\Cref{sec:norway} shows that \tessera{} improves wind-speed downscaling before any Norwegian observations are available. We investigate this cold-start advantage by asking whether held-out Norwegian stations have nearby analogues in the existing European training set under different location descriptors. We compare the descriptor spaces $\mathcal{D}$ used in \cref{sec:residual}: geography, elevation\,$+$\,mTPI ($\mathbf{e}$), the interpolated ERA5 static fields, the hand-crafted surface descriptor ($\mathbf{s}$), and \tessera{} ($\mathbf{z}_T$). Including $\mathbf{s}$ further probes \cref{sec:residual}'s findings that it is as informative as \tessera{} for persistent wind residual, asking whether the two representations are equally able to transfer this information to previously unseen locations.

Within the training set, the nearest-neighbor distance of each station is
\begin{equation}
  \nu^{\mathcal{D}}_s \;=\; \min_{s'\in\mathcal{T}(h),\, s'\neq s}
  \big\| d^{\mathcal{D}}_s - d^{\mathcal{D}}_{s'} \big\|_2 ,
  \qquad s\in\mathcal{T}(h),
  \label{eq:nn-dist}
\end{equation}
where $d_s^{\mathcal{D}}\in\mathbb{R}^{p_{\mathcal{D}}}$ denotes the feature vector in space $\mathcal{D}$ at station $s$. We take the reachability radius to be their $95$th percentile, $\tau^{\mathcal{D}}(h)=Q_{0.95}\big(\{\nu^{\mathcal{D}}_s : s\in\mathcal{T}(h)\}\big)$. A held-out station is reachable if it has a training neighbor within $\tau^{\mathcal{D}}$, and \emph{reachability} is the fraction of the out-of-training Norwegian set that has an in-distribution analogue in the training set,
\begin{equation}
  \mathrm{Reach}^{\mathcal{D}}(h) \;=\;
  \frac{1}{|\mathcal{H}^{\mathrm{unseen}}_{\mathrm{NO}}(h)|}
  \sum_{s\in\mathcal{H}^{\mathrm{unseen}}_{\mathrm{NO}}(h)}
  \mathbbm{1}\!\left[\,
    \min_{s'\in\mathcal{T}(h)} \big\| d^{\mathcal{D}}_s - d^{\mathcal{D}}_{s'} \big\|_2
    \;\le\; \tau^{\mathcal{D}}(h)
  \,\right].
  \label{eq:reach}
\end{equation}

High reachability alone does not establish that neighbors are informative analogues because a descriptor might omit characteristics that distinguish surface environments. We use \textit{separability} AUC as a control metric. For each descriptor, we fit an $L_2$-regularized logistic classifier under $4$-fold cross-validation to the membership label $z_s=\mathbbm{1}[\,s\in\mathcal{H}^{\mathrm{unseen}}_{\mathrm{NO}}(h)\,]$, giving each station an out-of-fold estimate $\hat p_s$. The resulting AUC is the probability the classifier ranks a random held-out Norwegian station above a random training station.

\begin{figure}[p]
  \centering
  \includegraphics[width=0.85\textwidth]{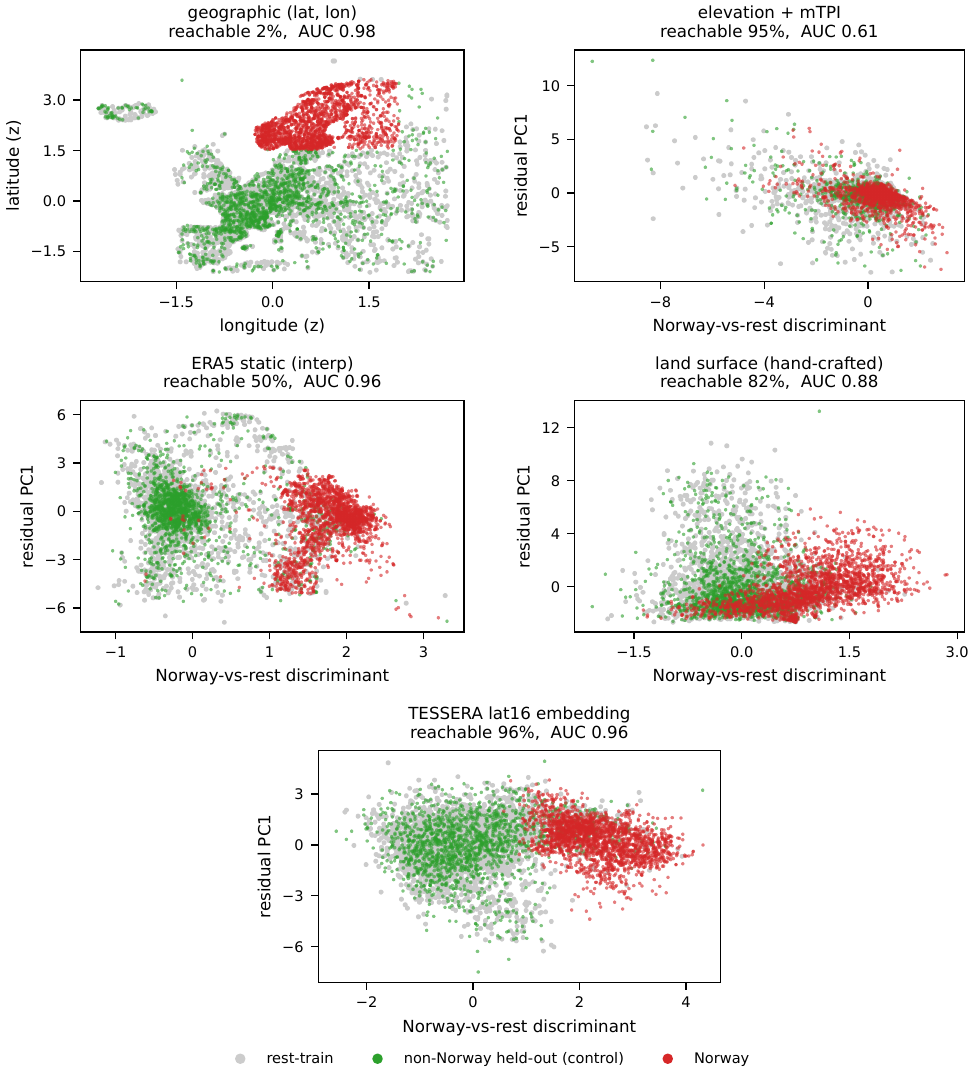}
  \caption{Where each descriptor places Norway relative to the European training set prior to deploying any stations. Each panel is a $2$-D view of one descriptor space. For non-geographic descriptor spaces, the horizontal axis is the \emph{Norway-vs-rest discriminant} --- each station's score along the linear combination of that descriptor's features that best separates Norway from the training set --- and the vertical axis is the leading principal component of the residual descriptor variation. The geographic panel shows standardised longitude and latitude. Grey: a subsample of the non-Norwegian European training stations; green: a non-Norwegian held-out control; red: Norway. In \tessera{} space, the Norwegian stations overlap locally with a subset of the European training stations, providing candidate surface analogues, while remaining globally distinguishable from the broader non-Norwegian distribution.}
  \label{fig:norway_descriptor_cluster}
\end{figure}

\paragraph*{\tessera{} provides both local support and descriptor fidelity at cold start.}
\Cref{fig:norway_descriptor_cluster} visualizes each descriptor space at the cold-start extreme. For each non-geographic descriptor, the horizontal axis is the normalized Norway-vs-rest discriminant: the leading direction $u^{\mathcal{D}}$ of a linear discriminant analysis separating Norwegian stations from non-Norwegian training stations in that descriptor space, normalized to unit length. A station's coordinate is the projection of its standardized descriptor onto this direction. The vertical axis carries the first principal component of what remains after this Norway-vs-rest direction is removed, $d^{\mathcal{D}}_s-\langle u^{\mathcal{D}}, d^{\mathcal{D}}_s\rangle\,u^{\mathcal{D}}$.

Geographically, Norway begins as an almost pure extrapolation problem, with only $\sim$$2\%$ of Norwegian stations having a nearby European training station. ERA5-static provides intermediate coverage of $\sim$$50\%$. Elevation\,$+$\,mTPI, by contrast, provides near-complete reachability at $95\%$. However, topography is a non-discriminative feature (AUC $\approx 0.6$): many European and Norwegian stations share similar elevation and terrain position while differing in surface characteristics. The hand-crafted surface descriptor does better at $0.88$ AUC, but leaves close to a fifth of the Norwegian stations without an in-distribution analogue ($82$\% reachability). 

\tessera{}, by contrast, combines near-complete local coverage at $96\%$ reachability with strong global separability of $0.96$ AUC --- its reachable neighbors are faithful surface analogues. This is illustrated by the red cluster overlapping locally with a subset of the grey European training stations while remaining strongly linearly separable from the non-Norwegian held-out stations in the green cluster.

\begin{figure}[t]
  \centering
  \includegraphics[width=0.85\textwidth]{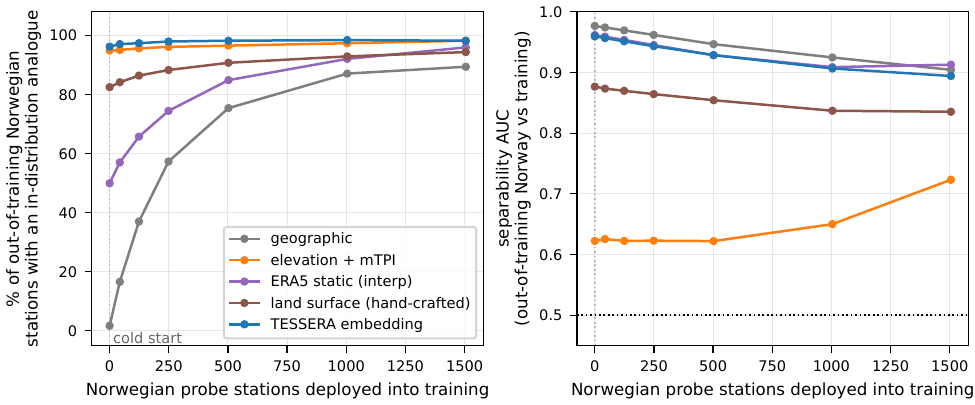}
  \caption{At each point along the phased deployment, the out-of-training Norwegian stations $\mathcal{H}^{\mathrm{unseen}}_{\mathrm{NO}}(h)$ (not-yet-deployed probes plus the permanently held-out Norwegian test set) are compared against the current training set $\mathcal{T}(h)$ (non-Norway Europe plus the probes deployed by horizon $h$) in five per-station descriptor spaces, all in standardized coordinates. \emph{(a)} Reachability; \emph{(b)} Separability AUC.}
  \label{fig:norway_reach_horizon}
\end{figure}

\Cref{fig:norway_reach_horizon} shows that the relative ordering of descriptor spaces changes little during deployment, indicating that cold-start comparisons capture the main descriptor-space geometry relevant to the experiment.

\paragraph*{Connection to the wind speed cold-start advantage.}
The analysis indicates \tessera{} provides the strongest combination of local coverage and descriptor fidelity, beating the hand-crafted surface descriptor $\mathbf{s}$ on both axes. This sharpens \cref{sec:residual}'s reading: the two spaces organize the persistent wind residual at unobserved stations, but they are not equally able to place an \emph{unobserved} station near an informative observed analogue. Therefore, the nearby, \textit{reachable} European training examples identified by \tessera{} are more likely relevant to the correction being transferred.

This provides an input-space explanation for \cref{sec:norway}'s results and clarifies why increasing station count alone does not remove \tessera{}'s advantage since, for wind speed, additional observations alone do not recover the missing surface structure information governing local correction. Nevertheless, this is a \emph{preliminary} geometric reading: reachability and separability describe how each feature clusters the stations, computed independently of the downscaling target. Whether that converts into downscaling skill depends on how much a location’s correction, per-variable, is governed by fine surface structure rather than by smooth, large-scale fields and the less discriminative elevation features both models share.

\bibliographystyle{ametsocV6}
\bibliography{refs}

\end{document}